\documentclass{article}

\usepackage{arxiv}

\usepackage[utf8]{inputenc} 
\usepackage[T1]{fontenc}    
\usepackage{hyperref}       
\usepackage{url}            
\usepackage{booktabs}       
\usepackage{amsfonts}       
\usepackage{nicefrac}       
\usepackage{microtype}      
\usepackage{lipsum}		    
\usepackage{graphicx}
\usepackage{natbib}
\usepackage{doi}
\usepackage{subfig}

\title{Explainable AI for Earth Observation: Current Methods, Open Challenges, and Opportunities}

\author{
\href{https://orcid.org/0000-0002-2294-4462}{\includegraphics[scale=0.06]{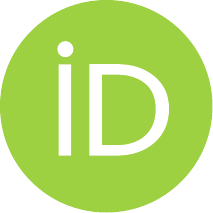}\hspace{1mm}
Gulsen~Taskin }
 \\
	Institute of Disaster Management\\
	Istanbul Technical University\\
    Istanbul, Turkiye \\
	\texttt{gulsen.taskin@itu.edu.tr} \\
	\And
	\href{https://orcid.org/0000-0001-6168-2883}{\includegraphics[scale=0.06]{orcid.pdf}\hspace{1mm}Erchan~Aptoula} \\
	Faculty of Engineering and Natural Sciences\\
	Sabanci University\\
	Istanbul, Turkiye \\
	\texttt{erchan.aptoula@sabanciuniv.edu} \\
	\AND
Alp~Ert\"urk 
\\
	Electronics and Telecom. Eng. Dept.\\
	Kocaeli University\\
	Kocaeli, Turkiye \\
	\texttt{alp.erturk@kocaeli.edu.tr} \\
}

\renewcommand{\shorttitle}{\textit{arXiv} Template}

\hypersetup{
pdftitle={A template for the arxiv style},
pdfsubject={q-bio.NC, q-bio.QM},
pdfauthor={Gulsen~Taskin, Erchan~Aptoula, Alp Erturk},
pdfkeywords={Deep learning,  Earth observation,  Explainability,   Interpretability,  Remote sensing, and XAI},
}

\begin{document}
\maketitle

\begin{abstract}
	Deep learning has taken by storm all fields involved in data analysis, including remote sensing for Earth observation. However, despite significant advances in terms of performance, its lack of explainability and interpretability, inherent to neural networks in general since their inception, remains a major source of criticism. Hence it comes as no surprise that the expansion of deep learning methods in remote sensing is being accompanied by increasingly intensive efforts oriented towards addressing this drawback through the exploration of a wide spectrum of Explainable Artificial Intelligence techniques. This chapter, organized according to prominent Earth observation application fields, presents a panorama of the state-of-the-art in explainable remote sensing image analysis.
\end{abstract}


\section{Introduction}\label{sec0}
The proliferation of airborne and space-borne imaging devices in the last few decades, combined with their ever-increasing spatial, spectral, and temporal resolutions, has led to publicly accessible, regularly acquired images of global coverage. This situation has exacerbated the already great need for efficient and effective data analysis tools. As the data acquisition rate has greatly surpassed the number of experts and their speed of knowledge provision, machine learning and data-oriented models have attracted considerable attention with respect to physical models for data analysis. Moreover, the advent of deep learning, along with its record-breaking performances across the computer vision and natural language processing landscape, has led to its widespread use in most, if not all, remote sensing-related tasks, including but not limited to scene classification, content-based retrieval, semantic segmentation, target detection, data fusion, change detection, and environmental monitoring \cite{Zhu2017}. The success of deep learning in a wide range of applications has resulted in a paradigm shift in pattern recognition, where data-driven approaches have de facto replaced traditional hand-crafted content descriptions used with legacy machine learning models and end-to-end systems, requiring primarily only an abundance of high-quality data.

\begin{figure}[ht]
\begin{center}
\includegraphics[width=0.8\textwidth]{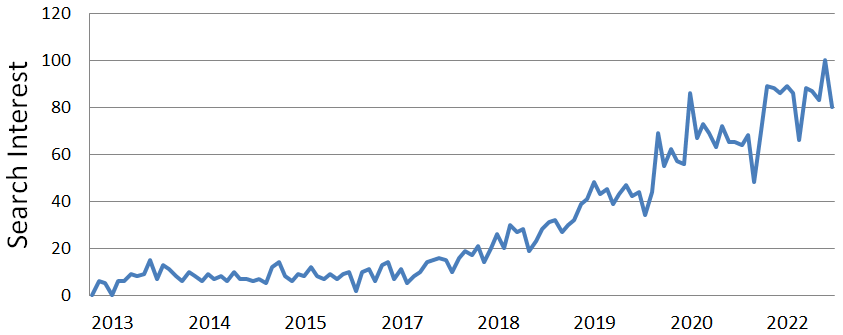} 
\end{center}
\caption{A plot of the search interest for the term ``Explainability'' during the last decade according to Google.}
\label{fig:searchPlot}
\end{figure}

Given the humble beginnings of artificial neural networks approximately 80 years ago \cite{mcculloughpitts}, the field has undoubtedly changed tremendously, first with the discovery of effective training algorithms \cite{rumelhart1986learning} and then through the catalytic effect of hardware acceleration, thus addressing one by one the initially long list of criticisms against artificial neural networks, such as the lack of effective training algorithms, the vanishing gradient problem, the lack of weight initialization strategies, etc. However, one source of criticism still remains: the lack of explainability and interpretability. Artificial neural networks, regardless of their specific type (e.g. multilayer perceptron, convolutional neural network, recurrent neural network, auto-encoder, vision transformer, etc.), are largely described as ``black-box'' methods with poor interpretability, where given a prediction, the cause-and-effect relations are challenging to establish. Therefore, developing more transparent and interpretable models is crucial to ensure the responsible use of machine learning in various real-world applications. One approach to addressing the lack of interpretability in black-box models is eXplainable Artificial Intelligence (XAI), providing a comprehensive understanding of the inner workings of machine learning models, thus enabling users to comprehend the rationale behind a particular decision or prediction. The last decade has seen a rapid increase in the number of published XAI studies (Fig.~\ref{fig:searchPlot}). 

The unprecedented spread of deep learning methods across almost all fields and sub-fields of remote sensing, coupled with the requirement of transparency and understandability in many Earth observation tasks, has further aggravated the lack of explainability inherent to deep learning models. This, in turn, has led to an increase in the number of published work on XAI for Earth observation, focused on rectifying this issue through a myriad of tools and techniques applied at all possible stages of data analysis, all claiming to improve in at least one way either the explainability and/or interpretability of deep learning based remote sensing image analysis methods. 

The present book chapter aims to provide an overview of XAI methods and approaches in the field of remote sensing so that the reader can quickly familiarize herself with the state-of-the-art of XAI in this field. Furthermore, despite the availability of recent surveys with similar goals, such as \cite{gevaert2022explainable}, the present work differentiates itself as it has been prepared specifically towards the remote sensing audience through a careful content organization that has been structured not according to XAI methodologies but instead according to primary remote sensing application areas.

\section{Research Methodology}\label{sec1}
\begin{figure}[ht]
\begin{center}
\includegraphics[width=0.9\textwidth]{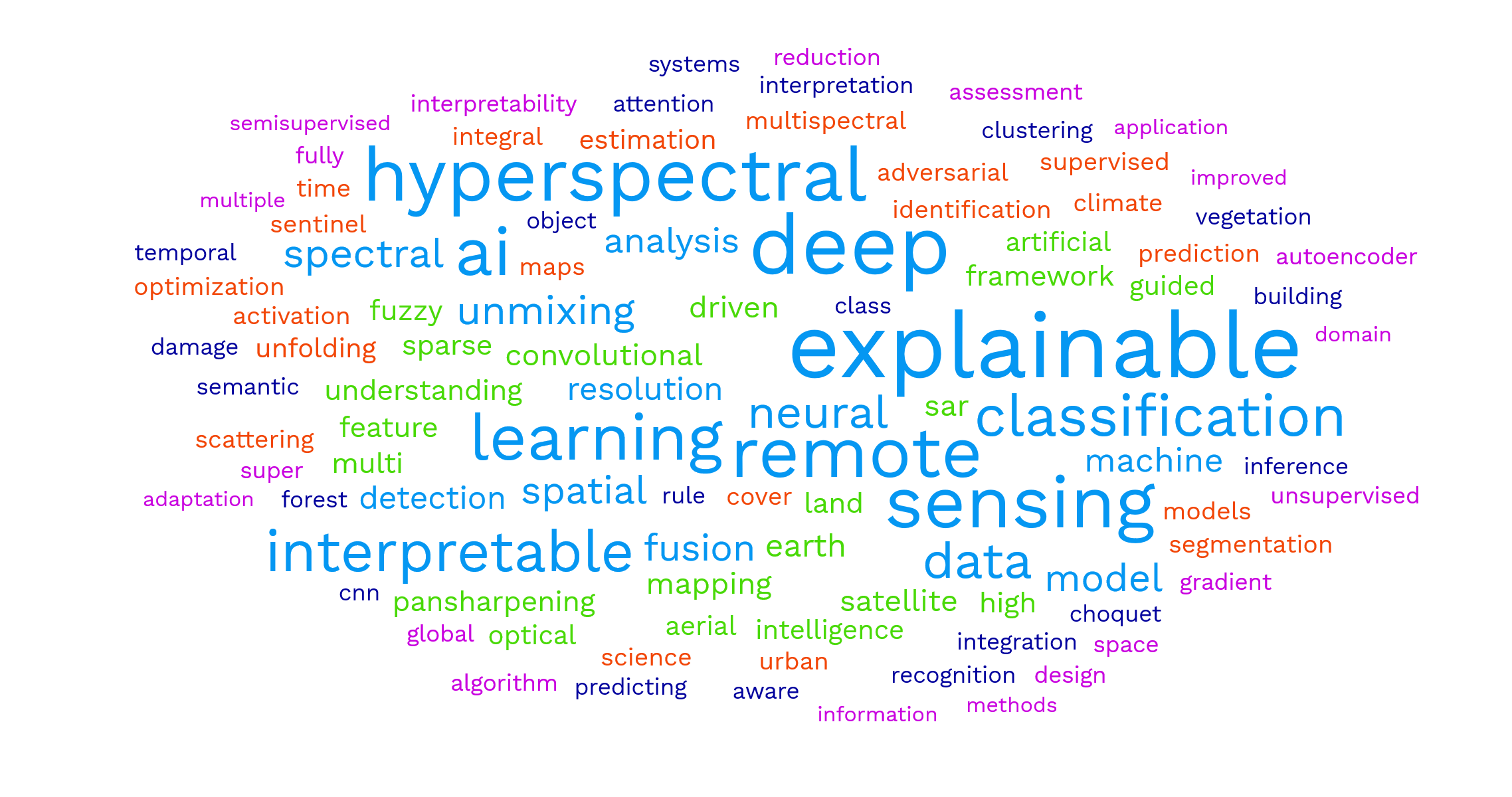} 
\end{center}
\caption{A world cloud showing the 100 most frequently encountered words in the titles of the scanned studies, sized according to their occurrence frequency.}
\label{fig:wordcloud}
\end{figure}

To conduct the state-of-the-art scan for XAI-related papers in the field of remote sensing, keywords such as \textit{explainable AI}, \textit{interpretable machine learning}, \textit{XAI}, \textit{model interpretation}, and \textit{interpretable deep learning}, were used in conjunction with \textit{remote sensing} as well as  remote sensing application areas such as \textit{classification}, \textit{target detection}, and \textit{fusion}. It was noticed that  \textit{interpretability} is a more commonly used keyword in remote sensing studies, although the keyword \textit{explainable} appears more frequently in the XAI nomenclature. Further examination revealed that not all studies using the keyword \textit{interpretability} were directly  related to XAI. After excluding such papers, only the publications related to the concept of explainable AI have remained. Figure~\ref{fig:wordcloud} represents a word cloud generated based on the collected publications related to explainable AI in the field of remote sensing, which form the basis of our database for this chapter. 

\section{Explainable Artificial Intelligence}\label{sec2}
In recent years, XAI has gained significant attention from researchers and a larger audience interested in understanding the cause-and-effect of machine learning models, as well as their internal structure and decision-making process. The goal is to improve the explainability and trustworthiness of the models, leading to a renewed focus on developing methods and algorithms that can render machine learning models more transparent and interpretable.  
Explainability has now become a requirement for various domains, including insurance risk assessment, data-driven medical diagnoses \cite{langlotz2019roadmap}, self-driving in autonomous vehicles, Earth system science \cite{reichstein2019deep}, and more, due to the importance of ensuring the reliability and compatibility of the model with the physical structure and processes of the real-world problems.

The roots for explainability can be traced back to earlier medical expert systems where efforts were made to explain decisions to physicians \cite{fagan1980computer,teach1981analysis,swartout1993explanation,ye1995impact}. The demand for explainability can also be seen in the insurance industry based on principles an decisions such as the US Equal Credit Opportunity Act Adverse Action Notice of 1974, which mandates the insurance companies by “\textit{providing statements of reasons in writing as a matter, of course, to applicants against whom adverse action is taken}” in the case when the customer’s application is rejected. In the present days, with the rise of the internet and the fast pace of machine learning, the European Parliament and Council have taken a big step towards legalizing the right to an explanation in the European Union General Data Protection Regulation (GDPR) framework \cite{gdpr}. Furthermore, the European Commission has formed a High-level Expert Group on Artificial Intelligence to shape the European Council’s strategy in AI. This group has issued the Ethics Guidelines for Trustworthy AI, which provides a general framework for the requirements of AI systems. As anticipated, the guideline emphasizes explainability as a top topic, obligating an AI system to provide an explanation on demand and “\textit{Such explanation should be timely and adapted to the expertise of the stakeholder concerned (e.g. layperson, regulator or researcher)}.” The need for explainability has also started to influence the industry under the influence of GDPR.  For example, the European Banking Federation recognizes the significance of transparency and explainability in AI systems used in the banking industry. Consequently, the industry is questioning the use of opaque black-box learning models \cite{Gade2020}. Explainability is not just a concern for GDPR; it is also a matter of interest to researchers and engineers who recognize the importance of causality in explanation \cite{pearl2009causal}. To explore this further, the Defense Advanced Research Projects Agency (DARPA) initiated a massive XAI program, forming eleven teams from universities and the military industry to investigate the topic from different perspectives, including a theory of explanation, evaluation frameworks, psychological requirements, human-computer interfaces, new post-hoc explanation techniques, and more \cite{gunning2019darpa}. 

All of these initiatives for XAI also promote the concept of responsible AI, incorporating the ethical and societal consequences of AI systems \cite{linardatos2020explainable,kusner2017counterfactual}. The principles of fairness, accountability, and transparency (FAT) are essential to ensure that AI systems accord with societal norms and are utilized responsibly and ethically. By offering transparency into the decision-making process of AI systems and making the internal structure of the model more interpretable and intelligible, XAI may be utilized to promote responsible AI that enables the discovery of any biases or discrimination in the system. 

New concepts, taxonomies, and definitions have been introduced with the emergence of the XAI field, some of which are specific to particular application areas, including remote sensing. Although review articles have attempted to explain and unify these concepts and definitions, their meanings may vary depending on the context in which they are used. In the following section, a brief overview of these definitions will be provided, emphasizing those that are particularly applicable to remote sensing.

\subsection{Taxonomy}
\begin{figure}[ht]
\begin{center}
\includegraphics[width=0.95\textwidth]{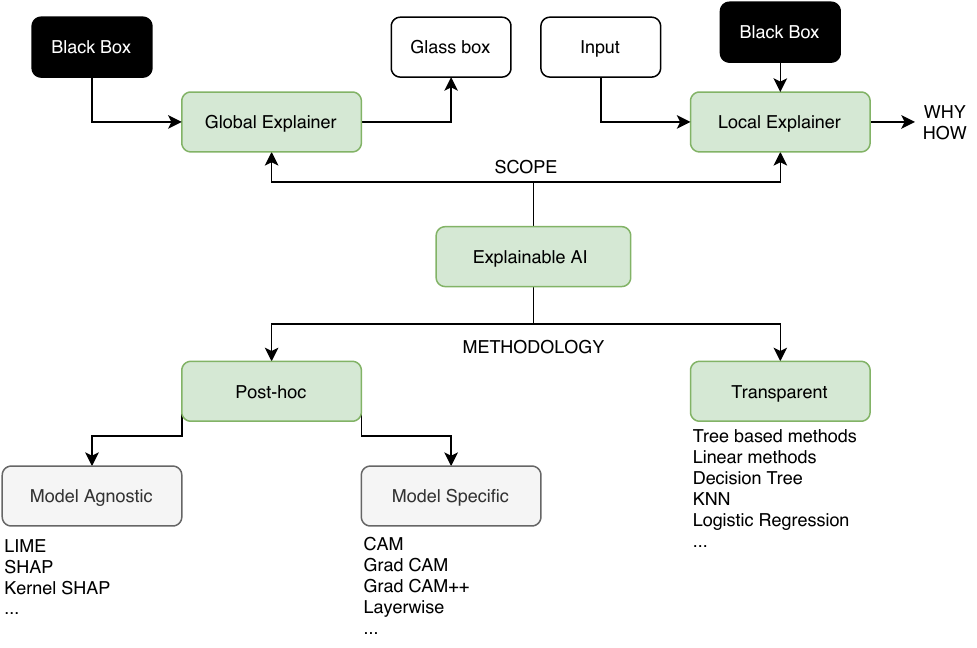} 
\end{center}
\caption{Illustration of XAI with respect to its scope and methodology.}
\label{fig:XAI_generalPlot}
\end{figure}

Explainability and interpretability are terms often used interchangeably in the field of XAI. However, although they have similar meanings, they are distinct concepts. Explainability refers to the ability of a model to clearly and accurately explain how and why it made a particular prediction, which involves presenting the reasoning behind a model's decisions in a way that humans easily understand. On the other hand, interpretability refers to the extent to which a model and its predictions can be understood by humans, which is associated with the ease or complexity of understanding a model's internal processes. Although interpretability is often associated with measuring the complexity of a model in the literature, it is frequently used as a synonym for comprehensibility.
Additionally, trustworthiness or  reliability is a concept used to evaluate the performance of machine learning models in terms of their consistency with domain knowledge. Transparency is another important concept in evaluating machine learning models as it enables humans to comprehend how it arrives at decisions and increases trust in its predictions. Besides these, there are many other concepts in the field of XAI. However, these concepts are generally not clearly distinguishable from each other, and some can be used together. In order to circumvent this issue, this work will employ the most widely accepted grouping method found in the literature. Accordingly, as shown in Figure \ref{fig:XAI_generalPlot}, it is possible to categorize all XAI methods into two main groups based on their scope and methodology. The scope refers to the level of explanation, which can be either global or local, while methodology refers to the approach or technique used to generate an explanation, which can be categorized as post-hoc or transparent. 

\subsection{Scopes of Explanations}
\textbf{Local methods} typically approximate the model’s behavior around a specific instance to be explained. Therefore, they may not necessarily generalize to the model's overall behavior. Local methods often perform by perturbing the values of each feature independently and measuring the change in the model's behavior using sensitivity analysis, gradients of the deep learning model, or feature scores, with the assumption that the model's behavior is approximately linear in the vicinity of the instance being explained \cite{fong2017}. It should be noted that these methods may not be effective when the model being explained is highly nonlinear, as the explanation may not accurately represent the model's behavior in other regions of the feature space. In image-based remote sensing applications, one of the most common techniques in this category is saliency maps, which is a visualization technique. A saliency map is a heatmap that highlights the salient parts of the input image (to be explained) that contributed most significantly to the model's prediction.

\textbf{Global methods} are used to provide an explanation of the model’s overall behavior; therefore, they are independent of the input samples to be explained. One of the well-known approaches in this category is model distillation, used to compress a large, complex model into a smaller, simpler, and, more importantly, interpretable one. These methods often operate by sampling synthetic training data across the entire input space in which the black-box model is evaluated. The distilled model is then trained using this synthetic data and a surrogate model, which globally mimics the behavior of the black-box model. The performance of global methods is strongly influenced by the size of the sampling space and the complexity of the surrogate model. Increasing the sampling space of training data may improve performance but increase computational cost, whereas increasing the complexity of the surrogate model may improve performance but decrease interpretability.  As it requires balancing the trade-off between model performance and interpretability, few studies aim to explain the black-box models globally. Moreover, it is essential to note that even with a large sample size, the accuracy of global approximation in image-based applications might still be low due to the curse of dimensionality caused by the enormous size of feature space.

\subsection{Types of Explanations}
XAI either provides an explanation for a particular black-box model, i.e. post-hoc explainability, or designs intrinsically transparent models, i.e. glass-box models \cite{Castelvecchi2016,Ribeiro2016}.

\subsubsection*{Transparent models} 
\begin{figure}[ht]
\begin{center}
\includegraphics[width=0.7\textwidth]{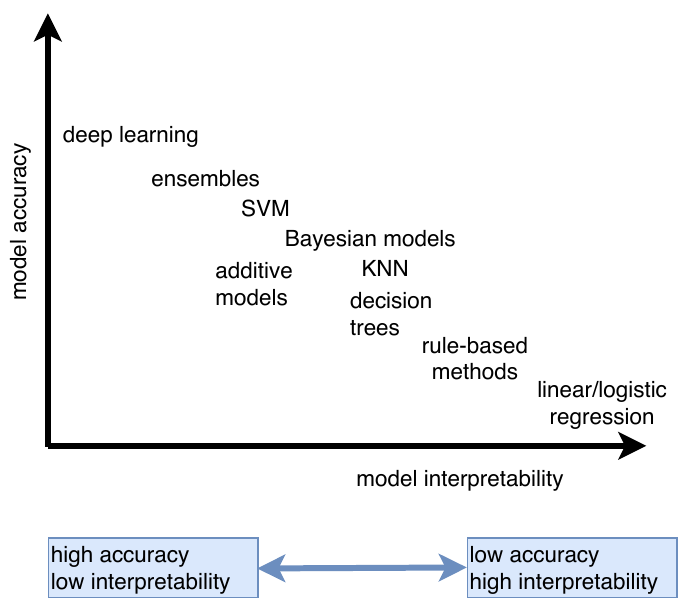}
\end{center}
\caption{A visualization of various machine learning methods in terms of the trade-off between interpretability and model performance.}
\label{fig:tradeoff}
\end{figure}

Transparent models provide a comprehensive understanding of interactions between all possible inputs and  how the learning model operates locally and globally. This enables traceability and the ability to discern the logic behind a specific prediction. For example, linear models, a common first choice, allow practitioners to trace model parameters to understand the internal process. Although they are the number one choice for many machine learning problems due to their simplicity,  their performance is limited for complex nonlinear problems. In order to address nonlinear problems, more complex transparent models can be designed, but this involves a trade-off between performance and interpretability, as illustrated in Figure~\ref{fig:tradeoff}. The XAI literature acknowledges that accuracy typically decreases as the interpretability of the learning model increases. Given this trade-off, most researchers focus on explaining black-box models rather than creating glass-box or transparent models. Examples of some transparent models are decision trees, logistic regression, and techniques originated based on generalized additive models \cite{hastie2017generalized}.

\subsubsection*{Post-hoc explainability} 
Post-hoc explainability aims to understand existing black-box models and explain how the decision model returned a specific outcome. This can be achieved with model distillation, explanation, and inspection. Model distillation aims to provide a global explanation of the whole logic of the learning model. In contrast, model explanation methods aim to locally understand the reasoning for the decision for a given input. On the other hand, the model inspection approaches use perturbation-based or sensitivity-based techniques to understand some specific properties of the interior behavior of the black box models when input is changed. The approaches for post-hoc explanation can be further categorized into model-agnostic and model-specific approaches. 

\textbf{Model-agnostic approaches} are used to explain the prediction of a machine learning model without relying on the specifics of the model being trained. Therefore, these approaches can be applied to any machine learning model, such as deep neural networks, regardless of its architecture. 

\textbf{Model-specific approaches} are tailored to the features and internal structure of a specific machine learning model, such as a specific deep learning model, random forest, or support vector machine. Although less versatile, these methods are generally more effective at explaining the model's predictions with respect to model-agnostic approaches because they are customized to the particular model used. 

\subsection{Methods for post-hoc explainability}
\subsubsection{Model-agnostic explanations}
There are various model-agnostic approaches in the literature, but feature importance scores, also referred to as feature selection methods, have been widely recognized and established for many years as one of the most commonly used model-agnostic approaches in several fields, including remote sensing. The more recent methods such as Local Interpretable Model-Agnostic Explanations (LIME) \cite{ribeiro2016should}, SHapley additive exPlanations (SHAP) \cite{lundberg2017unified}, and their variants have emerged, utilizing a similar concept of feature selection, albeit with differing approaches for determining the importance scores. However, this section will only focus on the commonly used model-agnostic techniques in XAI, instead of the feature selection methods that have been extensively studied in the past.

\subsubsection*{Local Interpretable Model-Agnostic Explanations (LIME)} 
LIME \cite{ribeiro2016should} aims to explain the prediction of a machine learning model in terms of a weighted linear regression. To achieve this, LIME starts with an instance to be explained and generates perturbations of the original instance by introducing random noise or perturbations to create a set of new instances. The machine-learning model then evaluates these perturbed instances, and a regression model is fitted to these local samples and their machine-learning outputs. The coefficients of this linear regression model are interpreted as feature importance scores to identify which feature has a more significant effect in determining the decision of the machine learning model. LIME may be utilized for text, tabular, and image inputs, but dimensionality needs to be reduced when an image is used as an input.  To reduce the dimensionality, a common approach is to group similar pixels into several smaller segments in which each segment is considered a feature. Then, LIME calculates the effect of each feature on the prediction of the machine-learning model to provide explanations. However, it should be noted that explanations only measure the effects of the segment on the model output and the behavior of the complex model in the local vicinity of the original instance to be explained. 

One of the drawbacks of LIME, particularly for large and complex models, is its computational requirements. To provide a local explanation, LIME needs to sample a large number of local instances, and each input that needs to be explained requires a separate regression model. This can be computationally expensive, especially for models with a large number of features or for large datasets. Another limitation of LIME is that it only captures linear correlations between features and the output of a machine learning model, as it relies on linear regression. Therefore, LIME may not be effective in explaining the behavior of non-linear models. In such cases, alternative explainability methods such as SHAP may be more suitable for providing interpretable explanations.

\subsubsection*{SHapley additive exPlanations (SHAP)}
SHAP \cite{lundberg2017unified} is another method that aims to explain the contribution of each input feature to a prediction made by a machine learning model. It is based on the Shapley value \cite{Shapley1953} from game theory, which is used to fairly distribute the payoffs from a game among all the players involved. In machine learning, the game is the prediction made by a model, the players are the input features that contribute to that prediction, and the payoff is the contribution of each feature to the prediction. 

SHAP is an additive method to provide an explainable approximated model, $f(x)$, for a complex original machine learning model, $g(x)$, as follows: 

\begin{equation}
f(x) = \phi_0 + \sum_{i=1}^n \phi_i x_i
\label{eq:shapadditive}
\end{equation}

where $\phi_0$ refers to the bias term, representing the average prediction of the machine learning model across all possible feature configurations, whereas $\phi_i$ refers to Shapley values for a specific $i-$th feature. According to Eq.~\ref{eq:shapadditive}, SHAP decomposes the prediction $f(x)$, into a sum of contributions from each feature, where the contributions of each feature are calculated using Shapley values. SHAP assigns a unique importance score to each feature in a prediction, considering the complex interactions between the features. The Shapley score for the $i$-th variable takes into account the impact of including it on its own as well as in combination with all other variables. Typically, Shapley values can be either positive or negative depending on how much the feature contributes to the prediction. Positive Shapley values indicate that the corresponding feature increases the value of the prediction, whereas negative Shapley values indicate a decrease in the value of the prediction. The magnitude of the Shapley value reveals how much the feature affects the prediction.

\begin{figure}[ht]
\begin{center}
\includegraphics[width=1\textwidth]{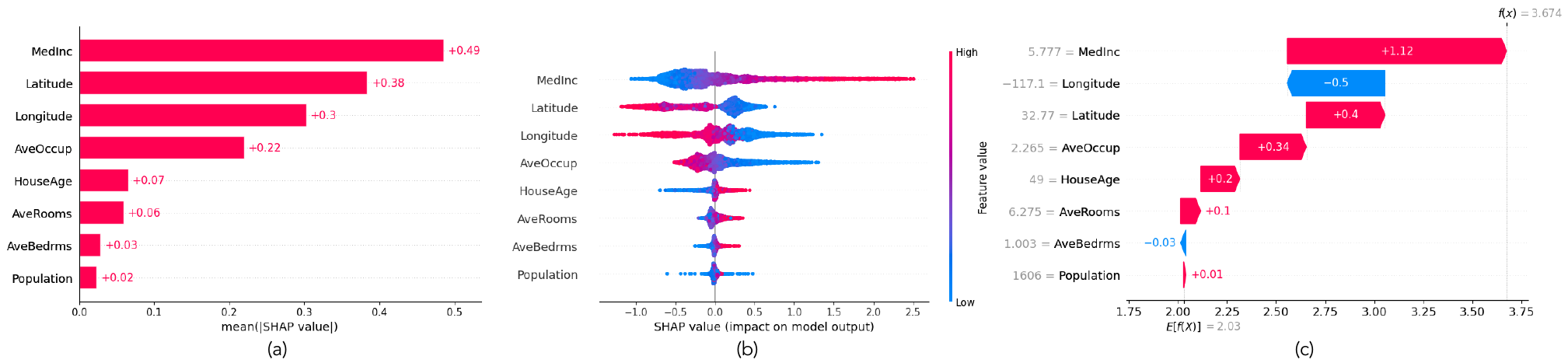}
\end{center}
\caption{A visualization of SHAP on California Housing dataset.}
\label{fig:shap}
\end{figure}

Figure \ref{fig:shap} illustrates the implementation of SHAP on the California Housing dataset. This dataset comprises eight distinct features, while the target variable (output) is the house pricing. In Figure \ref{fig:shap}(a), the mean SHAP values for each feature are depicted, representing their contributions to the model output. This can be considered a global explanation of the model. On the other hand, in Figure \ref{fig:shap}(b), individual samples are visualized as dots with varying colors between red and blue, reflecting their specific contributions to the model output. For instance, a higher value of the average number of bedrooms has a positive impact on house prices, while lower values have a negative impact. Furthermore, Fig. \ref{fig:shap}(c) demonstrates that SHAP also provides local explanations, revealing the contribution of each feature to the model output for a specific sample within the dataset.

It should be noted that as SHAP considers all possible interactions of the  features, its explanations are deemed more accurate and trustworthy and better equipped to take into account the full complexity of the model and its predictions. Unlike LIME, which only provides local explanations for individual model predictions, SHAP is considered a global and local method, providing a comprehensive understanding of the model and a detailed understanding of the contributions of individual features to the model's behavior.

\subsubsection*{Kernel SHAP}
SHAP is limited to machine learning models with an "additive" feature nature, where the effect of each feature can be independently decomposed. However, in most real-world cases, machine learning models are complex and nonlinear, making it difficult to use SHAP to explain them accurately. In order to address this limitation, Kernel SHAP \cite{lundberg2017unified}  was introduced as a model-independent approach that extends the classical LIME method. Unlike LIME, which uses heuristic definitions of locality, Kernel SHAP approximates feature contributions as Shapley values. An artificial dataset is required to train the Kernel SHAP model, where feature absence is simulated by replacing feature values with overall values from the training data. The method then trains a weighted linear regression model using artificial samples generated by turning features on or off. The resulting coefficients are used similar to Shapley values in order to explain the model's output.

\subsubsection{Model-specific explanations}
One commonly used technique in this category is gradient-based explanation methods, which are designed to explain the predictions made by deep neural networks by calculating the rate of change in the network output with respect to each feature. Model-specific explanation methods, such as integrated gradient \cite{Sundararajan2017} and DeepLIFT \cite{Shrikumar2016}, provide a linear approximation of a machine learning model's behavior that allows for determining the most essential features, also known as feature relevance. To accurately identify feature relevance, these methods must satisfy three axioms: sensitivity, completeness, and implementation invariance \cite{Sundararajan2017}. Integrated gradient \cite{Sundararajan2017}, which averages gradients of samples between the sample being explained and a chosen baseline, satisfies these axioms. DeepLIFT, on the other hand, compares the activation of a neuron to a reference activation to calculate feature relevance. Another method, Guided BackPropagation \cite{springenberg2014striving}, attempts to determine sensitivity by traversing from the output to the input layers in a deep convolutional neural network. These methods are often used to explain the behavior of complex black box models, such as deep neural networks \cite{Springenberg2015}. Other well-known methods for interpreting the decision-making process of convolutional neural networks (CNNs) include Class Activation Maps (CAM) \cite{zhou2016learning}, and Deep Taylor Decompositions \cite{montavon2018methods}. These methods generate saliency maps, highlighting significant parts of an image that contribute to the model's decision. However, these approaches can sometimes decrease the interpretability of the model due to linearization and discontinuity in gradient-based techniques. Although there are a lot more methods in computer vision applications, only those most frequently used in remote sensing will be described in this section.

\subsubsection*{Class Activation Mapping (CAM)}
Class Activation Mapping \cite{zhou2016learning} provides a heatmap that allows visualizing which parts of an input image are most important for a deep network's (such as CNN) prediction. In order to generate such a  heatmap, CAM maps the predicted class score back to the previous convolution layer.  Heatmap is created by taking the weighted sum of spectral channels of the last convolution layer, which is positioned right before global average pooling. The weights are selected to reflect the importance of each spectral channel in the class output. A common choice is to take the weights connecting the output of the pooling layer to the softmax layer. For multi-class problems, a different heatmap is generated for each class. The generated heatmaps also have a low spatial resolution. In order to visualize the heatmaps on the original input image, they are normalized and interpolated to the original input image size. 

\subsubsection*{Gradient-weighted Class Activation Mapping (Grad-CAM)}
Gradient-weighted class activation mapping (Grad-CAM) \cite{selvaraju2017grad} is a method for interpreting the decision-making process of a CNN. Similar to CAM, it generates a heat map with coarse-grained visualizations. In order to generate the heatmap, Grad-CAM feeds the gradients for a target class into the final convolutional layer and computes an importance score based on the gradients. It is worth mentioning that if an image contains multiple instances of the same object, Grad-CAM may not accurately highlight each \cite{huang2021better}.

\begin{figure}[htp]
\begin{center}
\subfloat[]{\label{fig.gradCam1}\includegraphics[width=0.25\textwidth]{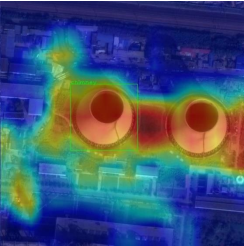}}\:
\subfloat[]{\label{fig.gradCam2}\includegraphics[width=0.25\textwidth]{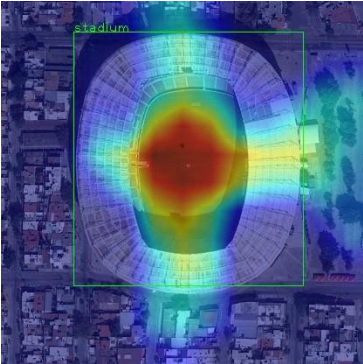}}\\
\subfloat[]{\label{fig.gradCam3}\includegraphics[width=0.25\textwidth]{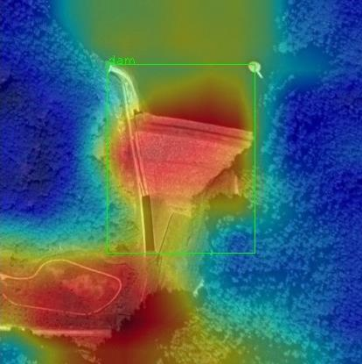}}\:
\subfloat[]{\label{fig.gradCam4}\includegraphics[width=0.25\textwidth]{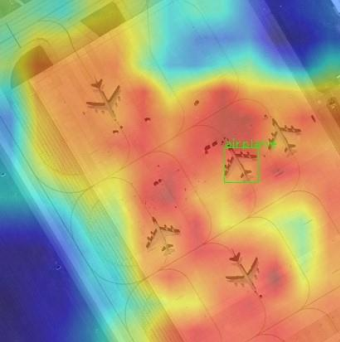}}\\
\subfloat[]{\label{fig.gradCam5}\includegraphics[width=0.25\textwidth]{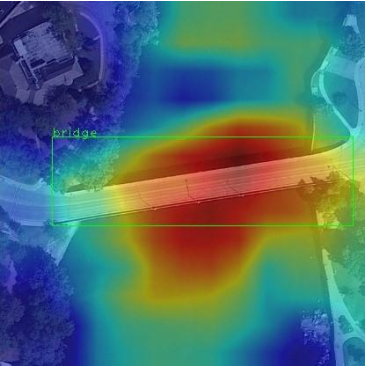}}\:
\subfloat[]{\label{fig.gradCam6}\includegraphics[width=0.25\textwidth]{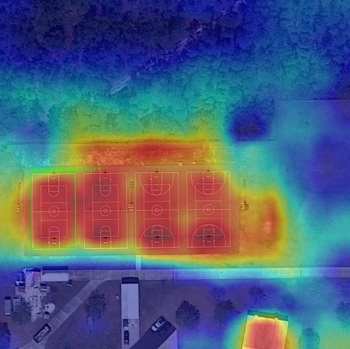}}
\end{center}
\caption{Grad-CAM heatmaps for target detection using YOLOv5s on the DIOR dataset} \label{gradcam_yolo}
\end{figure}

Fig.~\ref{gradcam_yolo} presents an application of Grad-CAM for a remote sensing application, specifically target detection. YOLOv5 target detection approach is trained on the DIOR dataset \cite{li2020object}, which contains optical images with 20 different types (or classes) of targets, and Grad-CAM is used in order to achieve interpretability in the target detection process. The highlighted regions in the heatmaps are those regions of the images that the deep network assigns higher weights, i.e.~more importance, in the detection process.

\subsubsection*{Grad-CAM++} 
Grad-CAM++ \cite{chattopadhay2018grad} was introduced to address the above limitations of Grad-CAM, considering the weighted average of the gradients to produce more fine-grained visualizations. In contrast to Grad-CAM, Grad-CAM++ provides a more detailed explanation for scenarios where multiple objects of the same category are present in a single image.

\section{Explainable AI in Remote Sensing}
Research on XAI in the fields of remote sensing and Earth sciences can be considered relatively limited \cite{maddy2021miidaps}. Many studies published in prestigious journals have pointed out the importance of XAI in various remote sensing applications, such as climate change monitoring and prediction, disaster management and response, agriculture and food security, urban planning, and development, etc., stating that black-box models, which tend to perform better than shallow learning models, do not necessarily meet the needs of remote sensing applications sufficiently \cite{lary2016machine, blair2019data}. This is because physical structures and processes are important to consider when making predictions or decisions in remote sensing. In addition, it has been emphasized that the underlying structures of complex models should be examined in detail due to several reasons, including improving the trustworthiness, accuracy, and interpretability of decisions \cite{roscher2020explainable,roscher2020explain,reichstein2019deep,camps2020advancing,hong2021interpretable,li2021advancing,karmakar2020feature}. 

The following sections analyze and present the status and state-of-the-art of the XAI in remote sensing literature.  The organization of these sections is not based on XAI approach groups but instead on significant remote sensing applications. The purpose of this organization is to provide a better vision to researchers in the fields of Earth observation and remote sensing.  In accordance with this organization, XAI in remote sensing has been investigated in eight subsections, namely pixel and scene classification, environmental monitoring, object (target) detection, unmixing, data fusion, synthetic aperture radar, multitemporal analysis, as well as miscellaneous applications.

\subsection{Pixel and Scene Classification}
Pixel-level image classification is one of the primary Earth observation data analysis tasks in the context of land-cover and land-use map production, whereas scene classification (and retrieval) is an emerging application stemming from the management need of ever-increasing data repositories. As such, they have both been extensively studied from an interpretability and explainability point of view.

One of the relatively early post-hoc interpretability attempts in the remote sensing classification context through deep learning has been realized by \citep{russwurm2018multi}, who proposed one of  the first approaches to approximate a phenological model for vegetation classes, via sequential recurrent encoders, based on Sentinel-2 images. In more detail, they visualize internal network activations over a sequence of cloudy and non-cloudy images, from which they deduce the network’s cloud filtering capacity. A significant post-hoc analysis study has been reported in \cite{campos2020understanding}, focusing on land use classification with Bidirectional Long Short Term Memory (BiLSTM), through Sentinel-2 time series acquired over Spain. The authors have explored feature importances and network activations to determine the most influential bands and deduced that increasing the number of layers does not necessarily lead to performance improvement. A similar post-hoc interpretation method has been employed in \cite{vasu2018aerial} with the aim of aerial scene classification from optical images through the popular UC Merced and AID datasets.  They have used class activation mapping techniques to visualize the deep network’s perception of aerial images and identify salient regions. Their study has shown that textures and local structures are of prime significance in this context.

CAM has been employed with the goal of interpreting CNNs in \cite{yang2019class} in the context of land cover mapping. Here, however, the approach is rather intrinsic since the class-specific saliency maps are used during training to quantify the contribution of the samples’ spatial regions and occlude the redundant areas, thus improving training performance. Another study relying on post-hoc interpretations has been reported in \cite{matrone2022bubblex}, where the authors propose a dynamic graph convolutional neural network-based framework for point-cloud classification. More specifically, they present both a visualization and interpretability module and employ Grad-CAM in order to produce coarse localization maps, highlighting important input regions. Thus, they claim to be able to identify the misleading features of misclassified objects.      

Grad-CAM has also been used more recently in \cite{qi2022embedded}, once again for post-hoc visualization purposes for interpretability. The authors present a multibranch, end-to-end trainable ensemble network for scene classification equipped with self-distillation so as to prune redundant ensemble branches during inference. Grad-CAM has been extended to 3D CNNs in \cite{de2022towards}, with the goal of hyperspectral remote sensing image classification for an edge computing environment. The authors have employed \textit{spectral-accumulation}, where a single value per pixel represents the activation for the selected class on all spectral bands. Recently, the study of \cite{huang2021better} reported that the Grad-CAM methods might fail in locating multiple target objects in a remote sensing image patch. In order to rectify this, they proposed a novel model called encoder-classifier-reconstruction CAM (ECR-CAM) neural network, which consists of four modules: an encoder module, a classifier module, a reconstruction module, and a CAM module. The reconstruction module is the key to locating more target objects by employing extracted features to reconstruct input images, allowing the features to retain important information about all objects. The CAM module shows more target objects with informative features, and the model improves the classification performance while accurately locating target objects.

\begin{figure}[htp]
\begin{center}
\subfloat[Asphalt]{\label{fig.shapAsphalt}\includegraphics[width=0.4\textwidth]{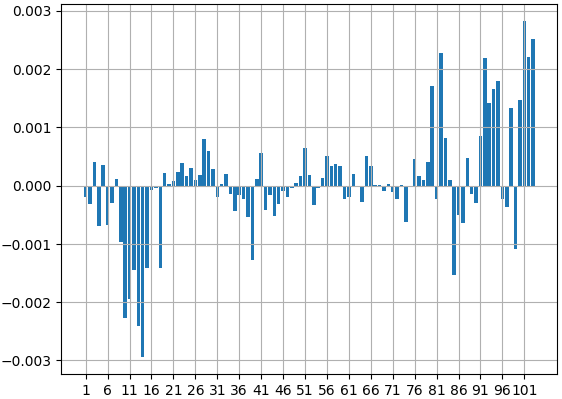}}\:
\subfloat[Meadows]{\label{fig.shapMeadows}\includegraphics[width=0.4\textwidth]{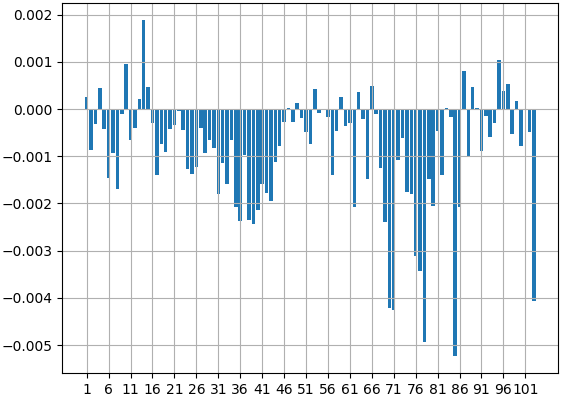}}\:
\subfloat[Gravel]{\label{fig.shapGravel}\includegraphics[width=0.4\textwidth]{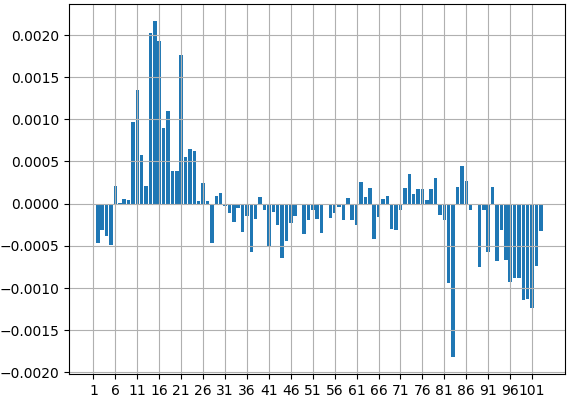}}\\
\end{center}
\caption{Band-based mean SHAP values for classes Asphalt,  Meadows, and Gravel, using a Random Forest Classifier on the Pavia University hyperspectral image.} \label{SHAP_Pavia1}
\end{figure}
\begin{figure}[htp]
\begin{center}
\subfloat[Trees]{\label{fig.shapTrees}\includegraphics[width=0.4\textwidth]{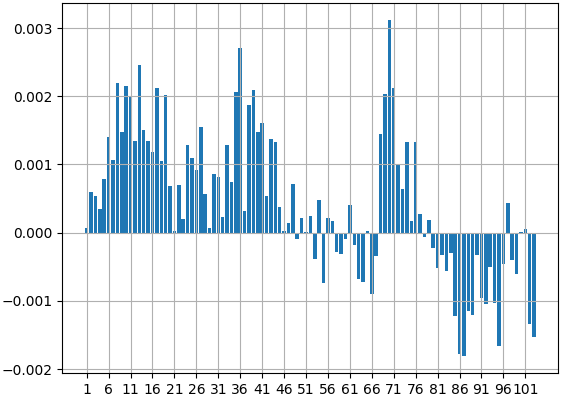}}\:
\subfloat[Metal Sheets]{\label{fig.shapSheets}\includegraphics[width=0.4\textwidth]{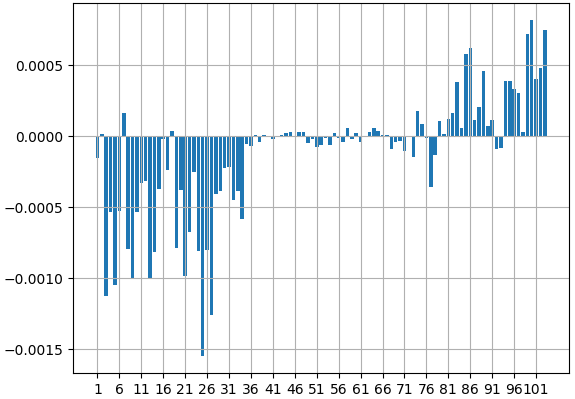}}\:
\subfloat[Soil]{\label{fig.shapSoil}\includegraphics[width=0.4\textwidth]{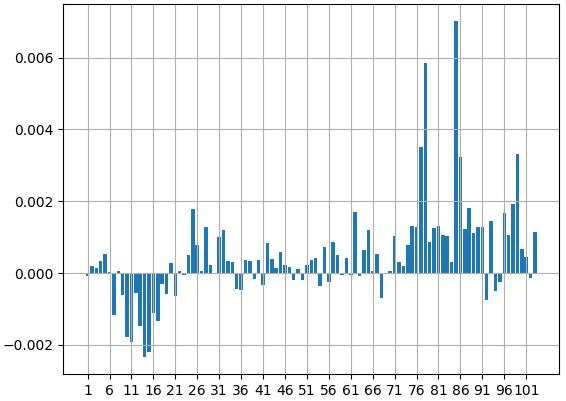}}    \\
\end{center}
\caption{Band-based mean SHAP values for classes Trees,  Metal Sheets, and  Soil, using a Random Forest Classifier on the Pavia University hyperspectral image.}
\label{SHAP_Pavia2}
\end{figure}
\begin{figure}[htp]
\begin{center}       
\subfloat[Bitumen]{\label{fig.shapBitumen}\includegraphics[width=0.4\textwidth]{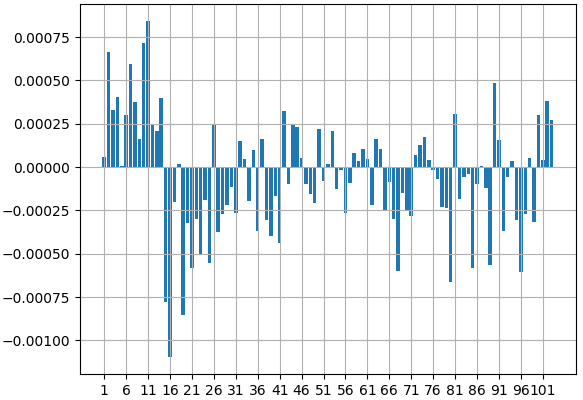}}\:
\subfloat[Bricks]{\label{fig.shapBricks}\includegraphics[width=0.4\textwidth]{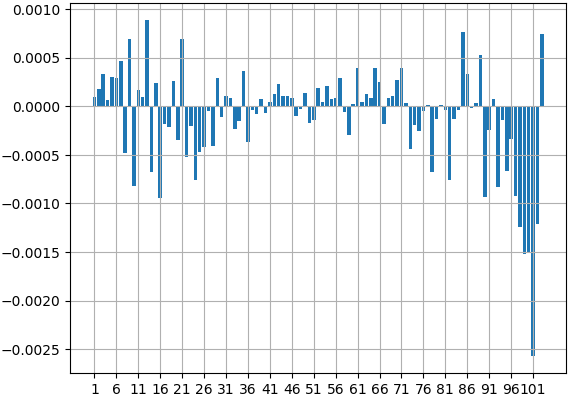}}\:
\subfloat[Shadows]{\label{fig.shapShadows}\includegraphics[width=0.4\textwidth]{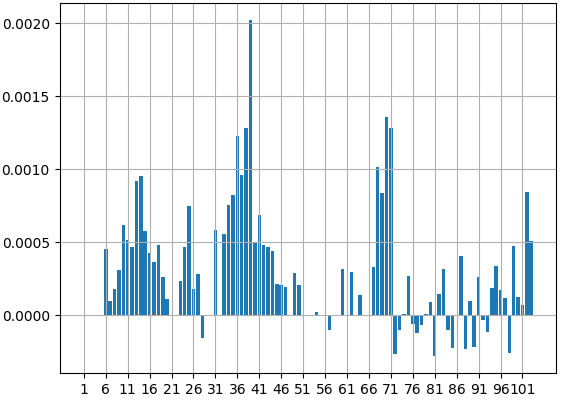}}    
\end{center}
\caption{Band-based mean SHAP values for classes Bitumen, Bricks and Shadows, using a Random Forest Classifier on the Pavia University hyperspectral image.}
\label{SHAP_Pavia3}
\end{figure}
\begin{figure}[h]
	\centering
    \includegraphics[scale=0.8]{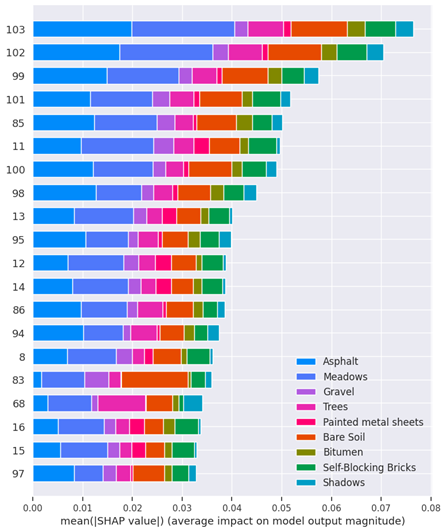}
	\caption{SHAP value plot for band-based interpretability, using Random Forest classifier on Pavia University hyperspectral image}
	\label{SHAPplotPavia}
\end{figure}

Another line of research along interpretability has focused on explaining and understanding individual predictions, commonly through SHAP and LIME. Examples of the former in remote sensing include \cite{matin2021earthquake}, where an MLP’s predictions have been analyzed for earthquake-induced building damage mapping via WorldView-3 and Open-Street-Map data  following the Palu, Indonesia earthquake in 2018. A further example of SHAP use is presented in \cite{abdollahi2021urban} for urban vegetation mapping from aerial imagery, where a deep neural network is provided with shallow features. LIME, on the other hand, has been utilized in \cite{verma2021explainable}, where it has been used to interpret CNN's predictions across the EuroSAT dataset. LIME has also been used in \cite{temenos2022novel}, where a heterogeneous spatio-temporal dataset from eight European cities acquired during the Covid-19 period is presented and analyzed in terms of monitoring the availability of public green spaces. Both LIME, SHAP, and its variation TreeSHAP are used in order to realize fast and accurate explanations \cite{temenos2022novel}. The results of a comprehensive XAI experiment series have been reported in \cite{kakogeorgiou2021evaluating}, in the context of multi-label scene classification via the BigEarth dataset.
Their results have shown Grad-CAM and LIME to be the most interpretable and reliable XAI methods, though computationally expensive. In \cite{fisher2022uncertainty}, SHAP is used to investigate how different features contribute to the model's predictions for slum mapping and is reported to indicate that a certain SWIR band is the most powerful feature for this task. SHAP is used for band-based interpretability of hyperspectral image classification in \cite{sahinbandbased}. Band-based mean SHAP values obtained using treeSHAP for each class when Random Forest classifier is trained on the Pavia University dataset \cite{sahinbandbased}. These class-based SHAP values are presented in Figs.~\ref{SHAP_Pavia1}, \ref{SHAP_Pavia2}, \ref{SHAP_Pavia3}. Fig.~\ref{SHAPplotPavia} presents the top twenty spectral bands with the highest SHAP mean values, and their respective importance for each class, in the Pavia University dataset. A recent study of \cite{temenos2023interpretable} proposes an XAI framework  for land use and land cover (LULC) classification in remote sensing using SHAP, allowing for both local and global explanations across different spectral bands. Their proposed approach considers different band combinations for classification and explanation, leading to improved accuracy and interpretability of results.


The relatively recently introduced deep rule-based classifiers have already been explored in the context of remote sensing. For instance, in \cite{gu2018massively}, the authors propose to use an ensemble of such classifiers for aerial scene classification, where each classifier is trained with a different level of spatial information. They report human-level performances through a transparent and parallelizable training process. A further example is reported in \cite{gu2018deep}, once again concentrating on scene classification and using a CNN as a feature descriptor, followed by a self-organizing set of transparent zero order fuzzy IF-THEN rules for classification. However, no comparison against the state-of-the-art is presented.

One of the representative examples of model distillation applied to remote sensing has been reported in \cite{guo2021network} for aerial scene classification. The authors rely on the interpretable CNN model \cite{zhang2018interpretable}, equipped with an extra loss for each filter in its convolutional layers so as to encourage the network towards object part representation. Thus such filters activate only with samples of certain categories. More specifically, the authors first train the interpretable CNN with a predefined pruning ratio and then proceed to remove the filters with poor interpretability. Another model distillation example has been provided in \cite{taskin2022model} for hyperspectral image classification. The author proposes a global model distillation approach to replace a black-box model with a fully explainable surrogate model utilizing polynomial chaos expansion.

In other studies, besides the aforementioned categories, the lack of interpretability among class labels in the context of multilabel image classification for high-resolution remote sensing images has been addressed in \cite{tan2022transformer}. The authors combine a semantic sensitivity module with a semantic relation-building module to generate content-aware class representations that are exploited through label relation inference. In addition, the authors of \cite{deshpande2021learning} tackle the interpretability of spectral features in terms of hyperspectral pixel classification. They propose the conversion of hyperspectral pixels into a spectral graph, followed by convolution. 

Scene and pixel-level classification in remote sensing constitute direct applications of machine learning to this field and unsurprisingly enjoy the attention brought by the advent of deep learning. The relatively intensive level of research and output level has also reflected the amount of effort towards explainability and interpretability. Most of them so far, however, has been rather direct applications of known techniques (CAM, Grad-CAM, SHAP, LIME, etc.) to this area, with little to no adaptation to the needs and particularities of remote sensing data.

\subsection{Environmental monitoring}
Environmental monitoring constitutes one of the primary application areas of remote sensing technologies, and its significance for contemporary societies is, in fact now greater than ever, as humanity’s environmental impact has started leading to changes at the global scale that require constant monitoring. As such, XAI is expected to be a crucial line of research on environmental monitoring in the near-future, with increasing concerns about accountability and transparency. 

Drought prediction has been explored in \cite{dikshit2021interpretable, dikshit2021explainable} via both LSTMs and Bi-LSTMs, and the spatial and temporal relationships between variables and prediction results were interpreted using the SHAP algorithm. The authors explored the inclusion of climatic variables (e.g., precipitation index) into the model, and their positive performance contribution was confirmed through the SHAP explainer. The aforementioned algorithm has also been employed in \cite{collini2022predicting} for interpreting the results of a series of methods, including autoencoders and CNNs, with the purpose of landslide prediction. 


Wildfire prediction, on the other hand, has been investigated in \cite{ronco2022explainable}, where the authors have employed a convolutional LSTM with ten years of data of spatio-temporal features and various weather indices. They have reported successful results through the combination of saliency maps with interpretable approximations such as LIME.

Continuing along the same disaster prediction theme, tropical cyclone prediction has been studied in \cite{xie2021visual}, where the authors propose two training strategies for a deep convolutional generative adversarial network so as to obtain prediction results with interpretable physical characteristics. These two strategies are long short-term training, and training a parameter selection according to physical characteristics. Disaster management has also been tackled in \cite{cheng2022uncertainty} for assessing post-disaster damage using aerial imaging. The authors have combined deep CNN-based multiclass classification and variational Bayesian inference in an effort to quantify uncertainty and improve model explainability for human decision-makers.  Earthquake-related remote sensing data analysis in the context of XAI has also been reported in the form of seismic faces analysis in \cite{li2020addcnn}. In an attempt to improve the geological or geophysical understandings of the relationships between the observations and background sciences, the authors have proposed a soft attention mechanism-based deep dilated convolutional neural network where subtle relations between the geological depositions and the seismic spectral responses are revealed by spatial-spectral attention maps.

An intrinsically interpretable method has been used in \cite{wang2022hyperspectral} for soil copper concentration estimation. The authors have based their approach on the attentive interpretable tabular learning model (TabNet), specially designed for tabular data. Besides its end-to-end trainability, TabNet employs sequential attention for feature selection and thus is reported to result in a more interpretable solution than its counterparts.

One of the few studies to tackle gold mine mapping has been reported in \cite{pradhan2022new}. The authors have used a convolutional neural network to estimate potential locations for gold mineralization in Eastern India. For the sake of interpretability, they have relied on SHAP values in order to determine the major contributing factors.

The use of XAI for crop monitoring and yield estimation has been reported widely. For example, in \cite{wolanin2020estimating}, both shallow and deep learning have been used for crop yield forecasting, and the yield drivers and features learned by the convolutional model have been visualized and analyzed through regression activation maps. The authors observed that images acquired during the crops’ growth season were the most effective in terms of forecasting. In another study \cite{perez2020interpretability}, addressing the same issue through multivariate time series, LSTMs have been used. The authors have employed three techniques with the goal of interpretability, namely: a permutation analysis of the input series, a qualitative visualization of activation maps, and their quantification via correlation analysis and clustering. In this way, they have determined redundant neurons. The rather specific case of grapevine classification was investigated in \cite{carneiro2022segmentation} to address the scarcity of grapevine variety professionals. The authors have used the eXception model along with a number of tools such as LIME, Grad-CAM, and Grad-CAM++ so as to visualize the segmentation impact in classification decisions. Their results have shown that their proposed approach focuses on more reliable regions for decision-making. Recently, the study of \cite{mateo2023interpretable} has developed LSTM architectures to estimate crop yields using multisensor satellite and meteorological data accurately and investigated the utility of interpreting the developed models using SHAP and integrated gradient (IG) techniques. Their results have shown that the proposed techniques effectively learned about crop phenology and yield, revealing the significance of several variables. 

As climate change is of paramount importance, it is not surprising that meteorology and climate science-oriented research equipped with XAI techniques has also been conducted. Notable and recent examples include \cite{mamalakis2022explainable}, where the authors present an overview of XAI applied to satellite applications such as weather phenomena identification and image-to-image translation, applications to climate prediction at sub-seasonal to decadal timescales, and detection of forced climatic changes and anthropogenic footprint. Estimating ground-level PM2.5 concentrations as an indicator of air quality has been explored in \cite{son2022sentinel}, where TabNet has been used, along with Sentinel-5P datasets and meteorological observations in order to estimate daily PM2.5 concentrations across Thailand. Another example appears in \cite{valdes2021machine}, where water vapor analysis has been conducted together with various XAI techniques like Permutational Variable Importance, LIME, Shapley Additive Explanations, and Ceteris Paribus profiles. Precipitation retrieval has been addressed in \cite{li2021advancing} via a Linknet segmentation followed by a tree ensemble. The authors have employed Grad-CAM so as to quantify the relative importance of spectral channels with respect to the rainfall identification problem. Recently, the study of \cite{stadtler2022explainable} discusses the relevance of air quality to human and environmental health and explains how explainable machine learning is used in air quality research. The authors used two different architectures, neural networks and random forests, trained on geospatial data to predict multi-year ozone averages. By analyzing inaccurate predictions and explaining why these predictions failed, they identified underrepresented samples and suggested new measurement stations, as well as determined which training samples were not essential to model performance. 

Furthermore, the global spatial suitability mapping of wind and solar systems has been explored in \cite{sachit2022global} via random forests, support vector machines, and multi-layer perceptrons, the predictions of which have been interpreted via the SHAP method.

Undoubtedly, environmental monitoring represents one of the remote sensing application areas with the most urgent need for reliable XAI in order to optimize the models’ ability to represent environmental events and phenomena. The results of the most exploratory studies conducted so far are highly promising and underline the great potential that XAI holds for the future.

\subsection{Object (Target) Detection}
Object or target detection is an important field of research in computer vision and remote sensing. In recent years, deep learning-based approaches, such as Fast Regional-based Convolutional Neural Networks (Fast R-CNN) \cite{girshick2015fast} and You-Only-Look-Once (YOLO) \cite{redmon2016you} have achieved high performances, particularly for high spatial resolution optical images. Unsurprisingly, exploration of XAI for target detection is also predominantly aimed towards optical images.  

In \cite{yang2019class}, Grad-CAM++ is used to generate the saliency maps of the input image, and the proposed class activation mapping guided adversarial training (CAMAT) architecture is built upon pre-trained YOLOv3 \cite{redmon2018yolov3} for target detection. Salient regions are then masked to improve the robustness and generalization performance of the network 
and the proposed methodology is validated in terms of detection performance on the NWPU VHR-10 dataset \cite{cheng2016learning}. In \cite{fu2019multicam}, fine-grained visual classification is introduced for aircraft recognition. The architecture consists of two networks, and multiple class activation mapping (MultiCAM) is proposed to locate the discriminative parts of objects from the target net, followed by a mask filter used to suppress background interference and features. In \cite{hogan2021towards}, Grad-CAM is used with YOLOv5 in order to obtain saliency maps for object detection in images acquired by unmanned aerial vehicles (UAVs) in the VisDrone dataset. In \cite{hogan2022explainable}, saliency maps of superpixels are generated by the perturbation-based KernelSHAP method for object detection in the VisDrone dataset by Yolov5. Furthermore, a bias is introduced to the dataset in order to evaluate the explainer’s ability to identify it, and pointing game \cite{zhang2018top} and deletion and insertion metrics \cite{petsiuk2018rise} are used to evaluate the explainer’s performance on single instances \cite{hogan2022explainable}. In \cite{kawauchi2022shap}, Gradient SHAP is used for feature attribution with YOLOv3 and Mask R-CNN \cite{he2017mask}, and COWC dataset \cite{mundhenk2016large} is used for performance evaluation. An evaluation metric is also proposed based on the feature attribution values \cite{kawauchi2022shap}. 


Other approaches for XAI in target detection include the critical feature capturing network (CFC-Net) \cite{ming2021cfc}, which utilizes a polarization attention module (PAM) to generate feature pyramids, a rotation anchor refinement module to refine anchors, and dynamic anchor learning , and which is validated in terms of detection performance on HRSC2016 \cite{liu2017high}, DOTA \cite{xia2018dota}, and UCAS-AOD \cite{zhu2015orientation} datasets. In \cite{xiong2022explainable}, the high-level feature maps obtained after a series of convolution and pooling operations in the convolutional backbone of VGG16 are fed into a causal multi-head attention model (CMAM) in order to obtain several attention maps. A filter aggregation mechanism (FAM) is used with the aim of making the convolutional filters more explainable \cite{xiong2022explainable}. FGSC-23 \cite{zhang2020new} and FGSCR-42 \cite{di2021public} datasets are used for performance evaluation of fine-grained ship classification, and an ablation study is used to support the use of CMAM \cite{xiong2022explainable}.   

Overall, the majority of XAI works in the field of object or target detection in remote sensing concerns the use of feature attribution methods such as Grad-CAM, SHAP, and their variants. Although the first works directly apply these methods for post-hoc interpretability of the target detection results, recent works use these methods as a way to guide the target detection network in terms of training data or network focus, for improved detection performance.

\subsection{Unmixing} \label{sec.unmixing}
Hyperspectral imaging sensors provide a wealth of spectral information, and the resulting high spectral resolution enables improved performance for many remote sensing tasks. However, the unavoidable low spatial resolution, particularly for the satellite-borne sensors, complex physical processes such as multiple scattering, and mixtures that inherently occur at the microscopic level, result in measured spectra being mixtures of the spectral signatures of the materials in the scene. Unmixing is the process of representing the pixel vectors of the data in terms of the spectral signatures of constituted spectra, assumed to be the pure materials in the scene, named endmembers, and their fractional abundances in each pixel \cite{keshava2002spectral}. Reviews on unmixing taxonomies, models, methods, challenges, and opportunities are available in \cite{bioucas2012hyperspectral, heylen2014review}.

Unmixing using physics-based mixture models is intrinsically interpretable, as the data is represented in terms of constituent material spectra and their fractional abundances. Before the recent interest in XAI and the prevalence of related taxonomy, this point was taken as given and was often left unstated. However, the recent interest in XAI resulted in an increase in the use of the term interpretability in unmixing literature. However, “\textit{interpretable results}” or “\textit{increased interpretability}” in a majority of papers in the literature still point to the inherent interpretability of the unmixing process and/or enhanced unmixing performance, and are therefore not in line with the scope of this Chapter.

Learning-based unmixing methods such as autoencoders, unlike physics-based mixture models, often are lacking in physical interpretability. A recent line of research focuses on providing interpretability for such methods by combining physics-based models with the deep learning approaches. In \cite{qian2020spectral}, linear mixture model and the iterative shrinkage-thresholding algorithm (ISTA) are unfolded to build two different schemes of network architecture, one for abundance estimation with a priori known endmembers, and one for blind unmixing. Experiments on synthetic and real datasets highlight that the approach outperforms regular deep learning-based unmixing approaches such as deep autoencoder networks (DAEN) \cite{qian2020spectral}. Xiong et al. have proposed a nonnegative matrix factorization inspired sparse autoencoder (NMF-SAE) approach, in which an L1-norm regularized NMF is paired with the deep learning network by uniting the weights across the layers in order to maintain the interpretability of physical models \cite{xiong2021nmf}. Hong et al. proposed a weakly-supervised unmixing network called WU-Net, which consists of a two-stream architecture, where one network is an autoencoder for unmixing, and the other network maps separately extracted endmembers to their abundances \cite{hong2019wu}. The two networks share the same weights in order to utilize the physics-based network’s physical interpretability \cite{hong2019wu}. In \cite{hong2021endmember}, this method is further improved to obtain the endmember-guided unmixing network (EGU-Net), in which the endmembers are extracted from partially overlapping blocks of the data, and clustering is used to aggregate endmembers into clusters, for the endmember network. A similar approach of a two-stream network is used in \cite{9489359}, in which one of the networks is still an autoencoder for unmixing, but the other uses endmember bundles extracted from superpixels of the scene with an autoencoder, and the two decoders share the weights. In \cite{xiong2021snmf}, $L_p, 0<p<1$ sparsity-constrained NMF is unrolled into an alternating deep network named SNMF-Net, which includes End-Net and Abun-Net submodules, for the endmembers and the abundances, respectively. Adversarial autoencoder network (AAENet) is proposed in \cite{jin2021adversarial}, which uses adversarial training to transfer spatial information into the network, which is derived and utilized via superpixels. 

A different line of research was investigated for interpretability in unmixing-based fusion of hyperspectral and panchromatic data in \cite{li2022unmixing}. The method proposed in \cite{li2022unmixing} injects high-frequency spatial details from the panchromatic image into the abundances and uses pixel-wise attention mechanisms for increased interpretability. 

Although the physical mixture models and the outputs of unmixing are interpretable by nature, deep learning-based unmixing methods may benefit from XAI. Currently, unmixing methods which aim to include interpretability for deep learning-based approaches almost exclusively adopt two-network approaches and autoencoders. Post-hoc interpretability and evaluation of interpretability are two subjects that are still not addressed in the unmixing literature. 

\subsection{Data Fusion}
Data fusion in remote sensing refers to the process of combining data from multiple sources, and may be used to provide increased information content, improved resolutions, increased robustness, and better exploitation of data. Data fusion in remote sensing is rapidly gaining importance due to advancements in sensor technology and versatility and the increasing amount of multi-scale and multi-resolution data being acquired. 

Data fusion for remote sensing has attracted significant interest in interpretability research. An important line of research in interpretability for data fusion is the use of unfolding into deep networks for pansharpening or multispectral (MS) and hyperspectral (HS) data fusion, and there have been a large number of works in the literature that utilize more or less similar strategies for combining model and data-driven approaches based on unfolding. Deep unfolding is the process of transferring an iterative algorithm into a deep learning architecture in order to derive, often a relatively small number of, trainable parameters. Unfolding enables to benefit from the performance of data-driven deep learning approaches for the optimization of parameters while retaining physical interpretability.  


In one of the earliest works for unfolding for data fusion, iterations of the projected gradient descent (PGD) algorithm are unrolled and replaced with a convolutional neural network (CNN) for pansharpening, and the proposed approach is evaluated on MS and panchromatic (PAN) images derived from AVIRIS hyperspectral images based on Wald protocol \cite{lohit2019unrolled}. Deep blind hyperspectral image fusion (DBIN), which uses Unfolding to CNN for HS and MS image fusion, instead of pansharpening, is proposed in \cite{wang2019deep}. DBIN is validated on CAVE \cite{yasuma2010generalized}, Harvard \cite{yasuma2010generalized}, and NTIRE2018 datasets. In \cite{xie2019multispectral}, a deep network for MS and HS image fusion named MHF-Net is proposed, which exploits the proximal gradient method and unfolding and adopts a deep residual network (ResNet) in order to learn the proximal operator. MHF-Net is evaluated on the CAVE dataset, data synthesized from the Hyperspec Chikusei dataset, and real Worldview-2 data, and is shown to outperform both traditional and deep learning-based fusion methods  \cite{xie2019multispectral}. This work is further improved in \cite{xie2020mhf}, which proposes two deep learning regimes for MHF-Net, namely consistent MHF-Net and blind MHF-Net, which are deemed more suitable when the training and test data are consistent and when there is a spectral or spatial mismatch between the training and test data, respectively. In \cite{xu2021deep}, a gradient projection-based pan-sharpening neural network (GPPNN), which unfolds the iterative steps into two networks and alternately stacks them in the backbone, is proposed. GPPN is evaluated on Landsat-8, Quickbird, and GaoFen-2 images and is shown to outperform MHF-Net, among other traditional and deep learning-based pan-sharpening methods \cite{xu2021deep}. Ablation experiments are conducted to prove that deep priors benefit the performance, and using separate networks for the MS and PAN blocks and not sharing the weights is shown to be the better approach \cite{xu2021deep}. The gradient descent algorithm is adopted, and the optimization process is unfolded for pansharpening in \cite{feng2021optimization}. Although unfolding terminology is not used in \cite{sun2021deep}, the proposed deep image prior-based interpretable network (DPIN) method operates under the same principle, and uses CNN for the iterative solution by half-splitting quadratic method. DPIN uses an encoding-decoding substructure to extract the deep image priors and uses these priors as a spatial guidance for the fusion \cite{sun2021deep}. DPIN is shown to outperform deep blind iterative fusion networks (DBIN). Coupled convolutional sparse coding-based pansharpening (PSCSC-Net) is proposed and evaluated on IKONOS, GeoEYE, and Worldview-2 data and is shown to outperform traditional and DL-based methods in \cite{yin2021pscsc}. In \cite{cao2021pancsc}, convolutional sparse coding is used with unfolding for the iterative algorithm in order to construct the proposed PanCSC-Net, and the method is evaluated on Worldview-III, Worldview-II, and GaoFen-2, and is shown to outperform traditional and deep-learning based pansharpening methods.  A deep unfolding method, LDUM, which includes a pre-treatment network consisting of multiple residual channel attention blocks (RCAB) and unfolding blocks,  is proposed for superresolution in \cite{wang2022deep}. LDUM is evaluated on the RRSSRD dataset, consisting of WorldView-II and GaoFen-2 images, and the AID dataset, consisting of Google Earth images, and is supported by ablation studies \cite{wang2022deep}. Image priors are learned via CNN-based residual learning in the hyperspectral sharpening method named DHSIS in \cite{dian2018deep}. In \cite{zheng2020hyperspectral}, a hyperspectral pansharpening method using deep hyperspectral priors and dual-attention residual network (DARN) is proposed. The approach uses channel attention and spatial attention modules to map the residual HS. Model guided deep convolutional network (MoG-DCN), which utilized deep denoiser and deep image priors with unfolding is proposed in \cite{dong2021model}, and is shown to outperform DBIN and MHF-Net. In \cite{shen2021admm}, for HS and MS fusion, the alternating direction method of multipliers (ADMM) is unrolled into CNN, and the SRF and point spread function (PSF) are learned adaptively in the proposed ADMM-HFNet approach. ADMM-HFNet is shown to provide compatible and often slightly better performances with respect to MHF-Net, DARN, and DBIN, whereas it has decreased computation time. Enhanced deep bling HIF network (EDBIN) proposed in \cite{wang2021enhanced} enforces bidirectional data consistency in the fusion process, and includes content-aware reassembly of features module \cite{wang2019carafe} and a SE-ResBlock \cite{hu2018squeeze} in order to lower distortion and redundancy. EDBIN is shown to outperform DBIN and ADMM-HFNet on CAVE and Harvard datasets. A model-guided unfolding network named DHIF-Net is proposed in \cite{huang2022deep}. DHIF-Net exploits a spatially adaptive 3D filter which is estimated from the HS-MS image pair by a deep network. Fusion is formulated as a differentiable optimization problem and unfolded into CNN \cite{huang2022deep}. DHIF-Net is shown to outperform MHF-Net and DBIN and DHSIS on CAVE and Harvard datasets.

Unfolding / unrolling with variational approaches for data fusion have also been researched in the literature. In \cite{li2021variation}, a variational pansharpening model which explores the similarity of MS and PAN images from the sparsity of nonlinear transforms is solved by unrolling the iterative shrinkage-thresholding algorithm and replacing it with CNNs. The Variation-Net is evaluated on MS and PAN images derived based on Wald protocol from Gaofen-1 data, and Quickbird images are used to evaluate its generalization capability \cite{li2021variation}. In \cite{lei2021interpretable}, two variational models describing the relationships of the high-resolution MS (HRMS) image with the MS and PAN images are alternately solved with a proximal gradient descent algorithm, which is unfolded and replaced by a CNN. The proposed method is evaluated on real Gaofen-2 and Worldview-2 images, and images derived based on Wald protocol \cite{lei2021interpretable}. In \cite{tian2021vp}, a deep network for variational pansharpening named VP-Net, which uses the similarity of the PAN image and the intensity of the MS image, and which incorporates a data fidelity term and unrolls the variable splitting method, is proposed. The proposed method is evaluated on GaoFen-2, GeoEye-1, Quickbird, and Worldview-2 images \cite{tian2021vp}. A variational method that unfolds gradient descent algorithm with spectral response function (SRF) guided CNNs to group spectral bands, and which utilizes channel attention module to embed parameter self-learning and a loss function based on L1 norm and SAM, is proposed for superresolution in \cite{he2021spectral}. HSR-Net is evaluated on CAVE and Sen2OHS datasets and supported by ablation studies \cite{he2021spectral}. A variational network for HS-MS fusion (VaFuNet), which represents data degradation and priors by deep networks, and unfolds half-quadratic splitting is proposed in \cite{yang2021variational}. A multiscale nonlocal attention module is also proposed and embedded into the deep prior network,
and the method is shown to outperform MHF-Net on various datasets. In \cite{yang2022memory}, memory-augmented deep conditional neural network (MDCUN), which includes two prior terms, namely denoising-based prior and non-local auto-regression prior (NARM), in the variational model is proposed. Proximal gradient projection is used for solving the sub-problems, and the iterative steps are unfolded to specific network modules, containing PAN-guided conditional band-ware MS denoise module, NARM, memory-augmented information module, and reconstruction module \cite{yang2022memory}. MDCUN is evaluated on WorldView-II, Worldview-III, and GaoFen-2 images, and is shown to outperform not only traditional and deep-learning-based methods but also a recent interpretable pan-sharpening approach, GPPNN. 


An interpretable spatial-spectral reconstruction network (SSR-NET) based on CNN, which does not involve unfolding but instead uses a model consisting of three components or modules, is proposed in \cite{zhang2020ssr} for HS and MS fusion. The first module is cross-mode message inserting (CMMI) which aims to pass information across the spatial and spectral modules; the second module is a spatial reconstruction network, and the third module is a spectral reconstruction network \cite{zhang2020ssr}. SSR-NET is modified in \cite{avagyan2022modified} to include long and short skip-connections, instead of skip-connections in only the spatial and spectral modules, and additional convolution and ReLU blocks in order to enhance feature extraction performance. MSSR-NET is shown to outperform SSR-NET for hyperspectral superresolution \cite{avagyan2022modified}. Another approach that does not involve unfolding is \cite{zhang2020deep} which proposes a deep generator network to capture the statistics of latent fused data, i.e., the SRF and PSF, based on the deep image priors \cite{lempitsky2018deep}.  

A different line of research, also mentioned in Section \ref{sec.unmixing}, is the unmixing-based interpretable PAN detail injection network proposed in \cite{li2022unmixing} for HS and PAN image fusion, particularly for relatively high spatial resolution ratios. The proposed method utilizes a pixel-wise attention mechanism and relates the PAN image with the abundances derived by unmixing \cite{li2022unmixing}. Comparable performances with state-of-the-art methods are obtained on various data synthesized from real data based on Wald’s protocol and a real dataset obtained by the ZY-1 02D satellite \cite{li2022unmixing}. 

A recent review of the coupling model and data-driven methods for image restoration and data fusion may be found in \cite{shen2022coupling}, which highlights the benefit of this avenue of reseach, and concludes that removing dependency on a large number of training samples and exploring proper data-driven priors based on an optimization-inspired variational mode remain significant challenges. 

The majority of works on XAI for data fusion in remote sensing focus on integrating model-based and deep learning-based approaches for increased pansharpening or HS-MS fusion performance. This integration is most often carried out by unfolding or unrolling optimization processes into deep learning. It should be noted that a significant number of these papers do not use XAI terminology or address interpretability. Therefore, as to be expected, evaluations are done for fusion performance, and no evaluation of interpretability is carried out.  

\subsection{SAR}
XAI has been relatively scarcely addressed in the context of synthetic aperture radar (SAR) image analysis. Notable examples include \cite{beker2022explainability}, where authors study the detection of long-term volcanic deformations via SAR interferometry, and using standard XAI tools such as Grad-CAM and t-SNE, they identify the slope-induced signal and salt lake patterns responsible for the model's misclassifications and model class separability. 

Single and dual polarimetric SAR images, on the other hand, have been investigated in \cite{zhao2019contrastive}, where the authors have attempted to learn a physically interpretable deep learning model directly from the original backscattered data. 

Furthermore, Sentinel-1 SAR data over four years and from three areas worldwide have been analyzed in \cite{amri2022offshore}, with the aim of developing a solution for oil slick segmentation and detection. 
The authors propose, in particular, an adaptation of the SHAP method to semantic segmentation in order to obtain visual explanations of deep learning predictions. 


Finally, a single study has been reported about synthetic aperture sonar imagery, in the context of seafloor texture classification \cite{walker2021explainable}, where LIME and divergence analysis are used in order to provide insight into the differences between baseline and fine-tuned deep fully connected networks. 

Overall, as far as SAR data is concerned, the exploration of XAI techniques can be so far quantified as exploratory at best.

\subsection{Multitemporal Analysis}
Multitemporal analysis of remote sensing data has gained prevalence in recent years, due to the increased number of EO satellites, improved temporal resolutions, and open data policies. However, XAI remains untapped for change detection and multitemporal analysis to a large extent.

A single work is included in this section, in which, in order to improve explainability, a convolutional autoencoder is deconstructed so that only those features selected based on variance are retained from the encoder layers, and new encoder-decoder layers are augmented iteratively 
\cite{bergamasco2020explainable}. The performance of the approach is validated qualitatively on only a single Landsat-8 bi-temporal image, but the premise of the approach is supported by the visualization of the retained features in \cite{bergamasco2020explainable}.


\subsection{Miscellaneous}
Besides the aforementioned categorized studies, additional reported investigations include an example of optical remote sensing image deblurring via \textit{deep unfolding} \cite{shi2022optical}, i.e.~a combination of traditional machine learning approaches with deep learning in the name of interpretability. 

A seminal position paper has also appeared \cite{tuia2021toward}, tackling the contemporary challenges of remote sensing, where interpretability and explainability are referenced as primary items in the agenda of Earth science data analysis.

Interpretability via semantic bottlenecks, in which predictions are related to human-interpretable attributes, has been explored for scenicness from the ScenicOrNot crowdsourcing database consisting of Sentinel-2 imagery in \cite{levering2020interpretable}. Semantic bottlenecks have also been explored for the investigation of housing factors and building quality scores in \cite{levering2021liveability}.    

\section{Discussion \& Conclusions}

 Remote sensing tasks and processes are often complex and opaque in nature. Although deep learning-based approaches provide significant advantages in terms of performance for many processing tasks, they are lacking in explainability and interpretability. XAI is crucial to overcome this drawback and can help to render the process more transparent, build trust, improve performances, validate results, and facilitate better decision-making. As such, it is no surprise that the need and interest in XAI in remote sensing are increasing. 
 
The state of XAI in remote sensing is still in its early stages, but there are some notable developments. The majority of remote sensing application areas have been explored in the context of XAI, albeit to different extents, and researchers have been exploring an increasing number of approaches for interpretability. However, the applications of XAI to remote sensing are still limited with a narrow subset of the available techniques, and most of the early studies have directly used the methods available from the computer vision field with almost no adaptation steps, disregarding the distinct nature of remote sensing data.

Currently, post-hoc interpretability techniques appear to constitute the majority of the implementations, particularly for the classification and detection tasks. However, the field has recently started to see an increase in methods that aim for intrinsic interpretability. Particularly for the fusion task, this has shown itself in approaches that aim to combine physics-based models with data-driven deep learning approaches. 

Based on the current status and the observed trends, the following remarks may be made about the future of XAI in remote sensing:

\begin{itemize}
\item The interest in XAI in remote sensing is bound to steadily increase, and an increasing number of XAI methods and approaches are due to appear in the remote sensing context.  
\item Remote sensing tasks and applications that have not yet significantly benefited from XAI approaches are also expected to see a relatively sharp increase in the number of works utilizing XAI in the following years. 
\item After the early direct applications of existing XAI methods in computer vision to remote sensing tasks, more elaborate adaptations of XAI methods and novel XAI approaches will be needed to address and account for the physics and particularities of remote sensing images.  
\item Explainability will need to be addressed not only for researchers but also for end users and policymakers in Earth observation sciences. 
\end{itemize}
\section*{Acknowledgement}
The first author would like to thank TUBITAK  for funding the research project under the grant number 122Y102. Erchan Aptoula was supported by the Sabanci University grant B.A.CF-23-02672.

\bibliographystyle{abbrvnat}
\bibliography{references}  

\begin{thebibliography}{151}
\providecommand{\natexlab}[1]{#1}
\providecommand{\url}[1]{\texttt{#1}}
\expandafter\ifx\csname urlstyle\endcsname\relax
  \providecommand{\doi}[1]{doi: #1}\else
  \providecommand{\doi}{doi: \begingroup \urlstyle{rm}\Url}\fi

\bibitem[Abdollahi and Pradhan(2021)]{abdollahi2021urban}
A.~Abdollahi and B.~Pradhan.
\newblock {Urban vegetation mapping from aerial imagery using Explainable AI
  (XAI)}.
\newblock \emph{Sensors}, 21\penalty0 (14):\penalty0 4738, 2021.

\bibitem[Amri et~al.(2022)Amri, Dardouillet, Benoit, Courteille, Bolon, Dubucq,
  and Credoz]{amri2022offshore}
E.~Amri, P.~Dardouillet, A.~Benoit, H.~Courteille, P.~Bolon, D.~Dubucq, and
  A.~Credoz.
\newblock Offshore oil slick detection: From photo-interpreter to explainable
  multi-modal deep learning models using sar images and contextual data.
\newblock \emph{Remote Sensing}, 14\penalty0 (15):\penalty0 3565, 2022.

\bibitem[Avagyan et~al.(2022)Avagyan, Katkovnik, and
  Egiazarian]{avagyan2022modified}
S.~Avagyan, V.~Katkovnik, and K.~Egiazarian.
\newblock Modified ssr-net: A shallow convolutional neural network for
  efficient hyperspectral image super-resolution.
\newblock \emph{Frontiers in remote sensing}, 2022.

\bibitem[Beker et~al.(2022)Beker, Ansari, Montazeri, Song, and
  Zhu]{beker2022explainability}
T.~Beker, H.~Ansari, S.~Montazeri, Q.~Song, and X.~X. Zhu.
\newblock Explainability analysis of {CNN} in detection of volcanic deformation
  signal.
\newblock In \emph{Proceedings of the IEEE International Geoscience and Remote
  Sensing Symposium (IGARSS)}, pages 4851--4854. IEEE, 2022.

\bibitem[Bergamasco et~al.(2020)Bergamasco, Saha, Bovolo, and
  Bruzzone]{bergamasco2020explainable}
L.~Bergamasco, S.~Saha, F.~Bovolo, and L.~Bruzzone.
\newblock An explainable convolutional autoencoder model for unsupervised
  change detection.
\newblock \emph{The International Archives of Photogrammetry, Remote Sensing
  and Spatial Information Sciences}, 43:\penalty0 1513--1519, 2020.

\bibitem[Bioucas-Dias et~al.(2012)Bioucas-Dias, Plaza, Dobigeon, Parente, Du,
  Gader, and Chanussot]{bioucas2012hyperspectral}
J.~M. Bioucas-Dias, A.~Plaza, N.~Dobigeon, M.~Parente, Q.~Du, P.~Gader, and
  J.~Chanussot.
\newblock Hyperspectral unmixing overview: Geometrical, statistical, and sparse
  regression-based approaches.
\newblock \emph{IEEE journal of selected topics in applied earth observations
  and remote sensing}, 5\penalty0 (2):\penalty0 354--379, 2012.

\bibitem[Blair et~al.(2019)Blair, Henrys, Leeson, Watkins, Eastoe, Jarvis, and
  Young]{blair2019data}
G.~S. Blair, P.~Henrys, A.~Leeson, J.~Watkins, E.~Eastoe, S.~Jarvis, and P.~J.
  Young.
\newblock Data science of the natural environment: a research roadmap.
\newblock \emph{Frontiers in Environmental Science}, 7:\penalty0 121, 2019.

\bibitem[Campos-Taberner et~al.(2020)Campos-Taberner, Garc{\'\i}a-Haro,
  Mart{\'\i}nez, Izquierdo-Verdiguier, Atzberger, Camps-Valls, and
  Gilabert]{campos2020understanding}
M.~Campos-Taberner, F.~J. Garc{\'\i}a-Haro, B.~Mart{\'\i}nez,
  E.~Izquierdo-Verdiguier, C.~Atzberger, G.~Camps-Valls, and M.~A. Gilabert.
\newblock Understanding deep learning in land use classification based on
  sentinel-2 time series.
\newblock \emph{Scientific reports}, 10\penalty0 (1):\penalty0 1--12, 2020.

\bibitem[Camps-Valls et~al.(2020)Camps-Valls, Reichstein, Zhu, and
  Tuia]{camps2020advancing}
G.~Camps-Valls, M.~Reichstein, X.~Zhu, and D.~Tuia.
\newblock Advancing deep learning for earth sciences: From hybrid modeling to
  interpretability.
\newblock In \emph{IGARSS 2020-2020 IEEE International Geoscience and Remote
  Sensing Symposium}, pages 3979--3982, 2020.

\bibitem[Cao et~al.(2021)Cao, Fu, Hong, Xu, and Meng]{cao2021pancsc}
X.~Cao, X.~Fu, D.~Hong, Z.~Xu, and D.~Meng.
\newblock Pancsc-net: A model-driven deep unfolding method for pansharpening.
\newblock \emph{IEEE Transactions on Geoscience and Remote Sensing},
  60:\penalty0 1--13, 2021.

\bibitem[Carneiro et~al.(2022)Carneiro, P{\'a}dua, Peres, Morais, Sousa, and
  Cunha]{carneiro2022segmentation}
G.~A. Carneiro, L.~P{\'a}dua, E.~Peres, R.~Morais, J.~J. Sousa, and A.~Cunha.
\newblock Segmentation as a preprocessing tool for automatic grapevine
  classification.
\newblock In \emph{Proceedings of the IEEE International Geoscience and Remote
  Sensing Symposium (IGARSS)}, pages 6053--6056. IEEE, 2022.

\bibitem[Castelvecchi(2016)]{Castelvecchi2016}
D.~Castelvecchi.
\newblock Can we open the black box of ai?
\newblock \emph{Nature News}, 538\penalty0 (7623):\penalty0 20, 2016.

\bibitem[Chattopadhay et~al.(2018)Chattopadhay, Sarkar, Howlader, and
  Balasubramanian]{chattopadhay2018grad}
A.~Chattopadhay, A.~Sarkar, P.~Howlader, and V.~N. Balasubramanian.
\newblock Grad-cam++: Generalized gradient-based visual explanations for deep
  convolutional networks.
\newblock In \emph{2018 IEEE winter conference on applications of computer
  vision (WACV)}, pages 839--847, 2018.

\bibitem[Cheng et~al.(2022)Cheng, Behzadan, and
  Noshadravan]{cheng2022uncertainty}
C.-S. Cheng, A.~H. Behzadan, and A.~Noshadravan.
\newblock Uncertainty-aware convolutional neural network for explainable
  artificial intelligence-assisted disaster damage assessment.
\newblock \emph{Structural Control and Health Monitoring}, 29\penalty0
  (10):\penalty0 e3019, 2022.

\bibitem[Cheng et~al.(2016)Cheng, Zhou, and Han]{cheng2016learning}
G.~Cheng, P.~Zhou, and J.~Han.
\newblock Learning rotation-invariant convolutional neural networks for object
  detection in vhr optical remote sensing images.
\newblock \emph{IEEE Transactions on Geoscience and Remote Sensing},
  54\penalty0 (12):\penalty0 7405--7415, 2016.

\bibitem[Collini et~al.(2022)Collini, Palesi, Nesi, Pantaleo, Nocentini, and
  Rosi]{collini2022predicting}
E.~Collini, L.~I. Palesi, P.~Nesi, G.~Pantaleo, N.~Nocentini, and A.~Rosi.
\newblock Predicting and understanding landslide events with explainable ai.
\newblock \emph{IEEE Access}, 10:\penalty0 31175--31189, 2022.

\bibitem[{De Lucia} et~al.(2022){De Lucia}, Lapegna, and Romano]{de2022towards}
G.~{De Lucia}, M.~Lapegna, and D.~Romano.
\newblock Towards explainable ai for hyperspectral image classification in edge
  computing environments.
\newblock \emph{Computers and Electrical Engineering}, 103:\penalty0 108381,
  2022.

\bibitem[Deshpande et~al.(2021)Deshpande, Thakur, and
  Balamuralidhar]{deshpande2021learning}
S.~Deshpande, R.~Thakur, and P.~Balamuralidhar.
\newblock Learning deep spectral features for hyperspectral data using
  convolution over spectral signature shape.
\newblock In \emph{2021 11th Workshop on Hyperspectral Imaging and Signal
  Processing: Evolution in Remote Sensing (WHISPERS)}, pages 1--5. IEEE, 2021.

\bibitem[Di et~al.(2021)Di, Jiang, and Zhang]{di2021public}
Y.~Di, Z.~Jiang, and H.~Zhang.
\newblock A public dataset for fine-grained ship classification in optical
  remote sensing images.
\newblock \emph{Remote Sensing}, 13\penalty0 (4):\penalty0 747, 2021.

\bibitem[Dian et~al.(2018)Dian, Li, Guo, and Fang]{dian2018deep}
R.~Dian, S.~Li, A.~Guo, and L.~Fang.
\newblock Deep hyperspectral image sharpening.
\newblock \emph{IEEE transactions on neural networks and learning systems},
  29\penalty0 (11):\penalty0 5345--5355, 2018.

\bibitem[Dikshit and Pradhan(2021{\natexlab{a}})]{dikshit2021explainable}
A.~Dikshit and B.~Pradhan.
\newblock {Explainable AI in drought forecasting}.
\newblock \emph{Machine Learning with Applications}, 6:\penalty0 100192,
  2021{\natexlab{a}}.

\bibitem[Dikshit and Pradhan(2021{\natexlab{b}})]{dikshit2021interpretable}
A.~Dikshit and B.~Pradhan.
\newblock {Interpretable and explainable AI (XAI) model for spatial drought
  prediction}.
\newblock \emph{Science of the Total Environment}, 801:\penalty0 149797,
  2021{\natexlab{b}}.

\bibitem[Dong et~al.(2021)Dong, Zhou, Wu, Wu, Shi, and Li]{dong2021model}
W.~Dong, C.~Zhou, F.~Wu, J.~Wu, G.~Shi, and X.~Li.
\newblock Model-guided deep hyperspectral image super-resolution.
\newblock \emph{IEEE Transactions on Image Processing}, 30:\penalty0
  5754--5768, 2021.

\bibitem[{European Parliament} and {Council of the European Union}(2016)]{gdpr}
{European Parliament} and {Council of the European Union}.
\newblock Regulation ({EU}) 2016/679 of the {European} {Parliament} and of the
  {Council} of 27 {April} 2016 on the protection of natural persons with regard
  to the processing of personal data and on the free movement of such data, and
  repealing {Directive} 95/46/{EC} ({General} {Data} {Protection}
  {Regulation}), 2016.

\bibitem[Fagan et~al.(1980)Fagan, Shortliffe, and Buchanan]{fagan1980computer}
L.~M. Fagan, E.~H. Shortliffe, and B.~G. Buchanan.
\newblock Computer-based medical decision making: from mycin to vm.
\newblock \emph{Automedica}, 3\penalty0 (2):\penalty0 97--108, 1980.

\bibitem[Feng et~al.(2021)Feng, Liu, Chen, Wang, and
  Zhao]{feng2021optimization}
Y.~Feng, J.~Liu, K.~Chen, B.~Wang, and Z.~Zhao.
\newblock Optimization algorithm unfolding deep networks of detail injection
  model for pansharpening.
\newblock \emph{IEEE Geoscience and Remote Sensing Letters}, 19:\penalty0 1--5,
  2021.

\bibitem[Fisher et~al.(2022)Fisher, Gibson, Liu, Abdar, Posa, Salimi-Khorshidi,
  Hassaine, Cai, Rahimi, and Mamouei]{fisher2022uncertainty}
T.~Fisher, H.~Gibson, Y.~Liu, M.~Abdar, M.~Posa, G.~Salimi-Khorshidi,
  A.~Hassaine, Y.~Cai, K.~Rahimi, and M.~Mamouei.
\newblock Uncertainty-aware interpretable deep learning for slum mapping and
  monitoring.
\newblock \emph{Remote Sensing}, 14\penalty0 (13):\penalty0 3072, 2022.

\bibitem[Fong and Vedaldi(2017)]{fong2017}
R.~C. Fong and A.~Vedaldi.
\newblock Interpretable explanations of black boxes by meaningful perturbation.
\newblock In \emph{Proceedings of the IEEE international conference on computer
  vision}, pages 3429--3437, 2017.

\bibitem[Fu et~al.(2019)Fu, Dai, Zhang, Wang, Yan, and Sun]{fu2019multicam}
K.~Fu, W.~Dai, Y.~Zhang, Z.~Wang, M.~Yan, and X.~Sun.
\newblock Multicam: Multiple class activation mapping for aircraft recognition
  in remote sensing images.
\newblock \emph{Remote sensing}, 11\penalty0 (5):\penalty0 544, 2019.

\bibitem[Gade et~al.(2019)Gade, Geyik, Kenthapadi, Mithal, and Taly]{Gade2020}
K.~Gade, S.~C. Geyik, K.~Kenthapadi, V.~Mithal, and A.~Taly.
\newblock Explainable ai in industry.
\newblock In \emph{Proceedings of the 25th ACM SIGKDD international conference
  on knowledge discovery \& data mining}, pages 3203--3204, 2019.

\bibitem[Gevaert(2022)]{gevaert2022explainable}
C.~M. Gevaert.
\newblock {Explainable AI for earth observation: A review including societal
  and regulatory perspectives}.
\newblock \emph{International Journal of Applied Earth Observation and
  Geoinformation}, 112:\penalty0 102869, 2022.

\bibitem[Girshick(2015)]{girshick2015fast}
R.~Girshick.
\newblock Fast {R-CNN}.
\newblock In \emph{Proceedings of the IEEE international conference on computer
  vision}, pages 1440--1448, 2015.

\bibitem[Gu and Angelov(2018)]{gu2018deep}
X.~Gu and P.~Angelov.
\newblock A deep rule-based approach for satellite scene image analysis.
\newblock In \emph{2018 IEEE International Conference on Systems, Man, and
  Cybernetics (SMC)}, pages 2778--2783. IEEE, 2018.

\bibitem[Gu et~al.(2018)Gu, Angelov, Zhang, and Atkinson]{gu2018massively}
X.~Gu, P.~P. Angelov, C.~Zhang, and P.~M. Atkinson.
\newblock A massively parallel deep rule-based ensemble classifier for remote
  sensing scenes.
\newblock \emph{IEEE Geoscience and Remote Sensing Letters}, 15\penalty0
  (3):\penalty0 345--349, 2018.

\bibitem[Gunning and Aha(2019)]{gunning2019darpa}
D.~Gunning and D.~Aha.
\newblock Darpa’s explainable artificial intelligence (xai) program.
\newblock \emph{AI magazine}, 40\penalty0 (2):\penalty0 44--58, 2019.

\bibitem[Guo et~al.(2021)Guo, Hou, Ren, Ren, and Jiao]{guo2021network}
X.~Guo, B.~Hou, B.~Ren, Z.~Ren, and L.~Jiao.
\newblock Network pruning for remote sensing images classification based on
  interpretable {CNN}s.
\newblock \emph{IEEE Transactions on Geoscience and Remote Sensing},
  60:\penalty0 1--15, 2021.

\bibitem[Hastie(2017)]{hastie2017generalized}
T.~J. Hastie.
\newblock Generalized additive models.
\newblock In \emph{Statistical models in S}, pages 249--307. Routledge, 2017.

\bibitem[He et~al.(2021)He, Li, Yuan, Shen, and Zhang]{he2021spectral}
J.~He, J.~Li, Q.~Yuan, H.~Shen, and L.~Zhang.
\newblock Spectral response function-guided deep optimization-driven network
  for spectral super-resolution.
\newblock \emph{IEEE Transactions on Neural Networks and Learning Systems},
  2021.

\bibitem[He et~al.(2017)He, Gkioxari, Doll{\'a}r, and Girshick]{he2017mask}
K.~He, G.~Gkioxari, P.~Doll{\'a}r, and R.~Girshick.
\newblock Mask {R-CNN}.
\newblock In \emph{Proceedings of the IEEE international conference on computer
  vision}, pages 2961--2969, 2017.

\bibitem[Heylen et~al.(2014)Heylen, Parente, and Gader]{heylen2014review}
R.~Heylen, M.~Parente, and P.~Gader.
\newblock A review of nonlinear hyperspectral unmixing methods.
\newblock \emph{IEEE Journal of Selected Topics in Applied Earth Observations
  and Remote Sensing}, 7\penalty0 (6):\penalty0 1844--1868, 2014.

\bibitem[Hogan and Aouf(2021)]{hogan2021towards}
M.~Hogan and N.~Aouf.
\newblock Towards real time interpretable object detection for uav platform by
  saliency maps.
\newblock In \emph{2021 IEEE International Conference on Robotics and
  Biomimetics (ROBIO)}, pages 1178--1183. IEEE, 2021.

\bibitem[Hogan et~al.(2022)Hogan, Aouf, Spencer, and
  Almond]{hogan2022explainable}
M.~Hogan, N.~Aouf, P.~Spencer, and J.~Almond.
\newblock Explainable object detection for uncrewed aerial vehicles using
  kernelshap.
\newblock In \emph{2022 IEEE International Conference on Autonomous Robot
  Systems and Competitions (ICARSC)}, pages 136--141. IEEE, 2022.

\bibitem[Hong et~al.(2019)Hong, Chanussot, Yokoya, Heiden, Heldens, and
  Zhu]{hong2019wu}
D.~Hong, J.~Chanussot, N.~Yokoya, U.~Heiden, W.~Heldens, and X.~X. Zhu.
\newblock Wu-net: A weakly-supervised unmixing network for remotely sensed
  hyperspectral imagery.
\newblock In \emph{Proceedings of the IEEE International Geoscience and Remote
  Sensing Symposium (IGARSS)}, pages 373--376. IEEE, 2019.

\bibitem[Hong et~al.(2021{\natexlab{a}})Hong, Gao, Yao, Yokoya, Chanussot,
  Heiden, and Zhang]{hong2021endmember}
D.~Hong, L.~Gao, J.~Yao, N.~Yokoya, J.~Chanussot, U.~Heiden, and B.~Zhang.
\newblock Endmember-guided unmixing network (egu-net): A general deep learning
  framework for self-supervised hyperspectral unmixing.
\newblock \emph{IEEE Transactions on Neural Networks and Learning Systems},
  2021{\natexlab{a}}.

\bibitem[Hong et~al.(2021{\natexlab{b}})Hong, He, Yokoya, Yao, Gao, Zhang,
  Chanussot, and Zhu]{hong2021interpretable}
D.~Hong, W.~He, N.~Yokoya, J.~Yao, L.~Gao, L.~Zhang, J.~Chanussot, and X.~Zhu.
\newblock Interpretable hyperspectral artificial intelligence: When nonconvex
  modeling meets hyperspectral remote sensing.
\newblock \emph{IEEE Geoscience and Remote Sensing Magazine}, 9\penalty0
  (2):\penalty0 52--87, 2021{\natexlab{b}}.

\bibitem[Hu et~al.(2018)Hu, Shen, and Sun]{hu2018squeeze}
J.~Hu, L.~Shen, and G.~Sun.
\newblock Squeeze-and-excitation networks.
\newblock In \emph{Proceedings of the IEEE conference on computer vision and
  pattern recognition}, pages 7132--7141, 2018.

\bibitem[Huang et~al.(2022)Huang, Dong, Wu, Li, Li, and Shi]{huang2022deep}
T.~Huang, W.~Dong, J.~Wu, L.~Li, X.~Li, and G.~Shi.
\newblock Deep hyperspectral image fusion network with iterative
  spatio-spectral regularization.
\newblock \emph{IEEE Transactions on Computational Imaging}, 8:\penalty0
  201--214, 2022.

\bibitem[Huang et~al.(2021)Huang, Sun, Feng, Ye, and Li]{huang2021better}
X.~Huang, Y.~Sun, S.~Feng, Y.~Ye, and X.~Li.
\newblock Better visual interpretation for remote sensing scene classification.
\newblock \emph{IEEE Geoscience and Remote Sensing Letters}, 19:\penalty0 1--5,
  2021.

\bibitem[Jin et~al.(2021)Jin, Ma, Fan, Huang, Mei, and Ma]{jin2021adversarial}
Q.~Jin, Y.~Ma, F.~Fan, J.~Huang, X.~Mei, and J.~Ma.
\newblock Adversarial autoencoder network for hyperspectral unmixing.
\newblock \emph{IEEE Transactions on Neural Networks and Learning Systems},
  2021.

\bibitem[Jin et~al.(2022)Jin, Ma, Mei, and Ma]{9489359}
Q.~Jin, Y.~Ma, X.~Mei, and J.~Ma.
\newblock Tanet: An unsupervised two-stream autoencoder network for
  hyperspectral unmixing.
\newblock \emph{IEEE Transactions on Geoscience and Remote Sensing},
  60:\penalty0 1--15, 2022.

\bibitem[Kakogeorgiou and Karantzalos(2021)]{kakogeorgiou2021evaluating}
I.~Kakogeorgiou and K.~Karantzalos.
\newblock Evaluating explainable artificial intelligence methods for
  multi-label deep learning classification tasks in remote sensing.
\newblock \emph{International Journal of Applied Earth Observation and
  Geoinformation}, 103:\penalty0 102520, 2021.

\bibitem[Karmakar et~al.(2020)Karmakar, Dumitru, Schwarz, and
  Datcu]{karmakar2020feature}
C.~Karmakar, C.~O. Dumitru, G.~Schwarz, and M.~Datcu.
\newblock Feature-free explainable data mining in sar images using latent
  dirichlet allocation.
\newblock \emph{IEEE Journal of Selected Topics in Applied Earth Observations
  and Remote Sensing}, 14:\penalty0 676--689, 2020.

\bibitem[Kawauchi and Fuse(2022)]{kawauchi2022shap}
H.~Kawauchi and T.~Fuse.
\newblock Shap-based interpretable object detection method for satellite
  imagery.
\newblock \emph{Remote Sensing}, 14\penalty0 (9):\penalty0 1970, 2022.

\bibitem[Keshava and Mustard(2002)]{keshava2002spectral}
N.~Keshava and J.~F. Mustard.
\newblock Spectral unmixing.
\newblock \emph{IEEE signal processing magazine}, 19\penalty0 (1):\penalty0
  44--57, 2002.

\bibitem[Kusner et~al.(2017)Kusner, Loftus, Russell, and
  Silva]{kusner2017counterfactual}
M.~J. Kusner, J.~Loftus, C.~Russell, and R.~Silva.
\newblock Counterfactual fairness.
\newblock \emph{Advances in neural information processing systems}, 30, 2017.

\bibitem[Langlotz et~al.(2019)Langlotz, Allen, Erickson, Kalpathy-Cramer,
  Bigelow, Cook, Flanders, Lungren, Mendelson, Rudie,
  et~al.]{langlotz2019roadmap}
C.~P. Langlotz, B.~Allen, B.~J. Erickson, J.~Kalpathy-Cramer, K.~Bigelow, T.~S.
  Cook, A.~E. Flanders, M.~P. Lungren, D.~S. Mendelson, J.~D. Rudie, et~al.
\newblock A roadmap for foundational research on artificial intelligence in
  medical imaging: from the 2018 nih/rsna/acr/the academy workshop.
\newblock \emph{Radiology}, 291\penalty0 (3):\penalty0 781, 2019.

\bibitem[Lary et~al.(2016)Lary, Alavi, Gandomi, and Walker]{lary2016machine}
D.~J. Lary, A.~H. Alavi, A.~H. Gandomi, and A.~L. Walker.
\newblock Machine learning in geosciences and remote sensing.
\newblock \emph{Geoscience Frontiers}, 7\penalty0 (1):\penalty0 3--10, 2016.

\bibitem[Lei et~al.(2021)Lei, Luo, Zhang, Li, Liu, and
  Li]{lei2021interpretable}
D.~Lei, X.~Luo, L.~Zhang, X.~Li, Q.~Liu, and W.~Li.
\newblock An interpretable deep neural network for panchromatic and
  multispectral image fusion.
\newblock In \emph{2021 7th International Conference on Big Data and
  Information Analytics (BigDIA)}, pages 71--78. IEEE, 2021.

\bibitem[Lempitsky et~al.(2018)Lempitsky, Vedaldi, and
  Ulyanov]{lempitsky2018deep}
V.~Lempitsky, A.~Vedaldi, and D.~Ulyanov.
\newblock Deep image prior.
\newblock In \emph{2018 IEEE/CVF Conference on Computer Vision and Pattern
  Recognition}, pages 9446--9454. IEEE, 2018.

\bibitem[Levering et~al.(2020)Levering, Marcos, Lobry, and
  Tuia]{levering2020interpretable}
A.~Levering, D.~Marcos, S.~Lobry, and D.~Tuia.
\newblock Interpretable scenicness from sentinel-2 imagery.
\newblock In \emph{IGARSS 2020-2020 IEEE International Geoscience and Remote
  Sensing Symposium}, pages 3938--3986. IEEE, 2020.

\bibitem[Levering et~al.(2021)Levering, Marcos, and
  Tuia]{levering2021liveability}
A.~Levering, D.~Marcos, and D.~Tuia.
\newblock Liveability from above: Understanding quality of life with overhead
  imagery and deep neural networks.
\newblock In \emph{2021 IEEE International Geoscience and Remote Sensing
  Symposium IGARSS}, pages 2094--2097. IEEE, 2021.

\bibitem[Li et~al.(2020{\natexlab{a}})Li, Zhou, Wang, and Wu]{li2020addcnn}
F.~Li, H.~Zhou, Z.~Wang, and X.~Wu.
\newblock Addcnn: An attention-based deep dilated convolutional neural network
  for seismic facies analysis with interpretable spatial--spectral maps.
\newblock \emph{IEEE Transactions on Geoscience and Remote Sensing},
  59\penalty0 (2):\penalty0 1733--1744, 2020{\natexlab{a}}.

\bibitem[Li et~al.(2020{\natexlab{b}})Li, Wan, Cheng, Meng, and
  Han]{li2020object}
K.~Li, G.~Wan, G.~Cheng, L.~Meng, and J.~Han.
\newblock Object detection in optical remote sensing images: A survey and a new
  benchmark.
\newblock \emph{ISPRS Journal of Photogrammetry and Remote Sensing},
  159:\penalty0 296--307, 2020{\natexlab{b}}.

\bibitem[Li et~al.(2021{\natexlab{a}})Li, Zhang, Tian, Ma, Zhou, and
  Wang]{li2021variation}
K.~Li, W.~Zhang, X.~Tian, J.~Ma, H.~Zhou, and Z.~Wang.
\newblock Variation-net: Interpretable variation-inspired deep network for
  pansharpening.
\newblock In \emph{2021 IEEE International Conference on Multimedia and Expo
  (ICME)}, pages 1--6. IEEE, 2021{\natexlab{a}}.

\bibitem[Li et~al.(2022)Li, Tian, Xia, and Liu]{li2022unmixing}
S.~Li, Y.~Tian, H.~Xia, and Q.~Liu.
\newblock Unmixing-based pan-guided fusion network for hyperspectral imagery.
\newblock \emph{IEEE Transactions on Geoscience and Remote Sensing},
  60:\penalty0 1--17, 2022.

\bibitem[Li et~al.(2021{\natexlab{b}})Li, Wen, Schreier, Behrangi, Hong, and
  Lambrigtsen]{li2021advancing}
Z.~Li, Y.~Wen, M.~Schreier, A.~Behrangi, Y.~Hong, and B.~Lambrigtsen.
\newblock Advancing satellite precipitation retrievals with data driven
  approaches: Is black box model explainable?
\newblock \emph{Earth and Space Science}, 8\penalty0 (2):\penalty0
  e2020EA001423, 2021{\natexlab{b}}.

\bibitem[Linardatos et~al.(2020)Linardatos, Papastefanopoulos, and
  Kotsiantis]{linardatos2020explainable}
P.~Linardatos, V.~Papastefanopoulos, and S.~Kotsiantis.
\newblock Explainable ai: A review of machine learning interpretability
  methods.
\newblock \emph{Entropy}, 23\penalty0 (1):\penalty0 18, 2020.

\bibitem[Liu et~al.(2017)Liu, Yuan, Weng, and Yang]{liu2017high}
Z.~Liu, L.~Yuan, L.~Weng, and Y.~Yang.
\newblock A high resolution optical satellite image dataset for ship
  recognition and some new baselines.
\newblock In \emph{International conference on pattern recognition applications
  and methods}, volume~2, pages 324--331. SciTePress, 2017.

\bibitem[Lohit et~al.(2019)Lohit, Liu, Mansour, and
  Boufounos]{lohit2019unrolled}
S.~Lohit, D.~Liu, H.~Mansour, and P.~T. Boufounos.
\newblock Unrolled projected gradient descent for multi-spectral image fusion.
\newblock In \emph{ICASSP 2019-2019 IEEE International Conference on Acoustics,
  Speech and Signal Processing (ICASSP)}, pages 7725--7729. IEEE, 2019.

\bibitem[Lundberg and Lee(2017)]{lundberg2017unified}
S.~M. Lundberg and S.-I. Lee.
\newblock A unified approach to interpreting model predictions.
\newblock \emph{Advances in neural information processing systems}, 30, 2017.

\bibitem[Maddy and Boukabara(2021)]{maddy2021miidaps}
E.~S. Maddy and S.~A. Boukabara.
\newblock Miidaps-ai: An explainable machine-learning algorithm for infrared
  and microwave remote sensing and data assimilation preprocessing-application
  to leo and geo sensors.
\newblock \emph{IEEE Journal of Selected Topics in Applied Earth Observations
  and Remote Sensing}, 14:\penalty0 8566--8576, 2021.

\bibitem[Mamalakis et~al.(2022)Mamalakis, Ebert-Uphoff, and
  Barnes]{mamalakis2022explainable}
A.~Mamalakis, I.~Ebert-Uphoff, and E.~A. Barnes.
\newblock {Explainable artificial intelligence in meteorology and climate
  science: Model fine-tuning, calibrating trust and learning new science}.
\newblock In \emph{International Workshop on Extending Explainable AI Beyond
  Deep Models and Classifiers}, pages 315--339, 2022.

\bibitem[Mateo-Sanchis et~al.(2023)Mateo-Sanchis, Adsuara, Piles,
  Munoz-Mar{\'\i}, Perez-Suay, and Camps-Valls]{mateo2023interpretable}
A.~Mateo-Sanchis, J.~E. Adsuara, M.~Piles, J.~Munoz-Mar{\'\i}, A.~Perez-Suay,
  and G.~Camps-Valls.
\newblock Interpretable long short-term memory networks for crop yield
  estimation.
\newblock \emph{IEEE Geoscience and Remote Sensing Letters}, 20:\penalty0 1--5,
  2023.

\bibitem[Matin and Pradhan(2021)]{matin2021earthquake}
S.~S. Matin and B.~Pradhan.
\newblock {Earthquake-induced building-damage mapping using Explainable AI
  (XAI)}.
\newblock \emph{Sensors}, 21\penalty0 (13):\penalty0 4489, 2021.

\bibitem[Matrone et~al.(2022)Matrone, Paolanti, Felicetti, Martini, and
  Pierdicca]{matrone2022bubblex}
F.~Matrone, M.~Paolanti, A.~Felicetti, M.~Martini, and R.~Pierdicca.
\newblock Bubblex: An explainable deep learning framework for point-cloud
  classification.
\newblock \emph{IEEE Journal of Selected Topics in Applied Earth Observations
  and Remote Sensing}, 15:\penalty0 6571--6587, 2022.

\bibitem[McCulloch and Pitts(1943)]{mcculloughpitts}
W.~S. McCulloch and W.~H. Pitts.
\newblock A logical calculus of the ideas immanent in nervous activity.
\newblock \emph{Bulletin of Mathematical Biophysics}, 5:\penalty0 115--133,
  1943.

\bibitem[Ming et~al.(2021)Ming, Miao, Zhou, and Dong]{ming2021cfc}
Q.~Ming, L.~Miao, Z.~Zhou, and Y.~Dong.
\newblock Cfc-net: A critical feature capturing network for arbitrary-oriented
  object detection in remote-sensing images.
\newblock \emph{IEEE Transactions on Geoscience and Remote Sensing},
  60:\penalty0 1--14, 2021.

\bibitem[Montavon et~al.(2018)Montavon, Samek, and
  M{\"u}ller]{montavon2018methods}
G.~Montavon, W.~Samek, and K.-R. M{\"u}ller.
\newblock Methods for interpreting and understanding deep neural networks.
\newblock \emph{Digital signal processing}, 73:\penalty0 1--15, 2018.

\bibitem[Mundhenk et~al.(2016)Mundhenk, Konjevod, Sakla, and
  Boakye]{mundhenk2016large}
T.~N. Mundhenk, G.~Konjevod, W.~A. Sakla, and K.~Boakye.
\newblock A large contextual dataset for classification, detection and counting
  of cars with deep learning.
\newblock In \emph{European conference on computer vision}, pages 785--800.
  Springer, 2016.

\bibitem[Pearl(2009)]{pearl2009causal}
J.~Pearl.
\newblock Causal inference in statistics: An overview.
\newblock \emph{Statistics surveys}, 3:\penalty0 96--146, 2009.

\bibitem[P{\'e}rez-Suay et~al.(2020)P{\'e}rez-Suay, Adsuara, Piles,
  Mart{\'\i}nez-Ferrer, D{\'\i}az, Moreno-Mart{\'\i}nez, and
  Camps-Valls]{perez2020interpretability}
A.~P{\'e}rez-Suay, J.~E. Adsuara, M.~Piles, L.~Mart{\'\i}nez-Ferrer,
  E.~D{\'\i}az, A.~Moreno-Mart{\'\i}nez, and G.~Camps-Valls.
\newblock Interpretability of recurrent neural networks in remote sensing.
\newblock In \emph{Proceedings of the IEEE International Geoscience and Remote
  Sensing Symposium (IGARSS)}, pages 3991--3994. IEEE, 2020.

\bibitem[Petsiuk et~al.(2018)Petsiuk, Das, and Saenko]{petsiuk2018rise}
V.~Petsiuk, A.~Das, and K.~Saenko.
\newblock Rise: Randomized input sampling for explanation of black-box models.
\newblock \emph{arXiv preprint arXiv:1806.07421}, 2018.

\bibitem[Pradhan et~al.(2022)Pradhan, Jena, Talukdar, Mohanty, Sahu, Raul, and
  Abdul~Maulud]{pradhan2022new}
B.~Pradhan, R.~Jena, D.~Talukdar, M.~Mohanty, B.~K. Sahu, A.~K. Raul, and K.~N.
  Abdul~Maulud.
\newblock A new method to evaluate gold mineralisation-potential mapping using
  deep learning and an explainable artificial intelligence (xai) model.
\newblock \emph{Remote Sensing}, 14\penalty0 (18):\penalty0 4486, 2022.

\bibitem[Qian et~al.(2020)Qian, Xiong, Qian, and Zhou]{qian2020spectral}
Y.~Qian, F.~Xiong, Q.~Qian, and J.~Zhou.
\newblock Spectral mixture model inspired network architectures for
  hyperspectral unmixing.
\newblock \emph{IEEE Transactions on Geoscience and Remote Sensing},
  58\penalty0 (10):\penalty0 7418--7434, 2020.

\bibitem[Redmon and Farhadi(2018)]{redmon2018yolov3}
J.~Redmon and A.~Farhadi.
\newblock Yolov3: An incremental improvement.
\newblock \emph{arXiv preprint arXiv:1804.02767}, 2018.

\bibitem[Redmon et~al.(2016)Redmon, Divvala, Girshick, and
  Farhadi]{redmon2016you}
J.~Redmon, S.~Divvala, R.~Girshick, and A.~Farhadi.
\newblock You only look once: Unified, real-time object detection.
\newblock In \emph{Proceedings of the IEEE conference on computer vision and
  pattern recognition}, pages 779--788, 2016.

\bibitem[Reichstein et~al.(2019)Reichstein, Camps-Valls, Stevens, Jung,
  Denzler, Carvalhais, et~al.]{reichstein2019deep}
M.~Reichstein, G.~Camps-Valls, B.~Stevens, M.~Jung, J.~Denzler, N.~Carvalhais,
  et~al.
\newblock Deep learning and process understanding for data-driven earth system
  science.
\newblock \emph{Nature}, 566\penalty0 (7743):\penalty0 195--204, 2019.

\bibitem[Ribeiro et~al.(2016{\natexlab{a}})Ribeiro, Singh, and
  Guestrin]{Ribeiro2016}
M.~T. Ribeiro, S.~Singh, and C.~Guestrin.
\newblock Model-agnostic interpretability of machine learning.
\newblock \emph{arXiv preprint arXiv:1606.05386}, 2016{\natexlab{a}}.

\bibitem[Ribeiro et~al.(2016{\natexlab{b}})Ribeiro, Singh, and
  Guestrin]{ribeiro2016should}
M.~T. Ribeiro, S.~Singh, and C.~Guestrin.
\newblock " why should i trust you?" explaining the predictions of any
  classifier.
\newblock In \emph{Proceedings of the 22nd ACM SIGKDD international conference
  on knowledge discovery and data mining}, pages 1135--1144,
  2016{\natexlab{b}}.

\bibitem[Ronco et~al.(2022)Ronco, Prapas, Kondylatos, Papoutsis, Camps-Valls,
  Fern{\'{a}}ndez-Torres, {Piles Guillem}, and
  Carvalhais]{ronco2022explainable}
M.~Ronco, I.~Prapas, S.~Kondylatos, I.~Papoutsis, G.~Camps-Valls, M.-{\'{A}}.
  Fern{\'{a}}ndez-Torres, M.~{Piles Guillem}, and N.~Carvalhais.
\newblock {Explainable deep learning for wildfire danger estimation}.
\newblock In \emph{EGU General Assembly Conference Abstracts}, pages
  EGU22----11787, 2022.

\bibitem[Roscher et~al.(2020{\natexlab{a}})Roscher, Bohn, Duarte, and
  Garcke]{roscher2020explain}
R.~Roscher, B.~Bohn, M.~F. Duarte, and J.~Garcke.
\newblock {Explain It to Me--Facing Remote Sensing Challenges in the Bio-and
  Geosciences With Explainable Machine Learning}.
\newblock \emph{ISPRS Annals of the Photogrammetry, Remote Sensing and Spatial
  Information Sciences}, 3:\penalty0 817--824, 2020{\natexlab{a}}.

\bibitem[Roscher et~al.(2020{\natexlab{b}})Roscher, Bohn, Duarte, and
  Garcke]{roscher2020explainable}
R.~Roscher, B.~Bohn, M.~F. Duarte, and J.~Garcke.
\newblock Explainable machine learning for scientific insights and discoveries.
\newblock \emph{IEEE Access}, 8:\penalty0 42200--42216, 2020{\natexlab{b}}.

\bibitem[Rumelhart et~al.(1986)Rumelhart, Hinton, and
  Williams]{rumelhart1986learning}
D.~E. Rumelhart, G.~E. Hinton, and R.~J. Williams.
\newblock Learning representations by back-propagating errors.
\newblock \emph{nature}, 323\penalty0 (6088):\penalty0 533--536, 1986.

\bibitem[Ru{\ss}wurm and K{\"o}rner(2018)]{russwurm2018multi}
M.~Ru{\ss}wurm and M.~K{\"o}rner.
\newblock Multi-temporal land cover classification with sequential recurrent
  encoders.
\newblock \emph{ISPRS International Journal of Geo-Information}, 7\penalty0
  (4):\penalty0 129, 2018.

\bibitem[Sachit et~al.(2022)Sachit, Shafri, Abdullah, Rafie, and
  Gibril]{sachit2022global}
M.~S. Sachit, H.~Z.~M. Shafri, A.~F. Abdullah, A.~S.~M. Rafie, and M.~B.~A.
  Gibril.
\newblock {Global Spatial Suitability Mapping of Wind and Solar Systems Using
  an Explainable AI-Based Approach}.
\newblock \emph{ISPRS International Journal of Geo-Information}, 11\penalty0
  (8):\penalty0 422, 2022.

\bibitem[Sahin et~al.(2023)Sahin, Erturk, and Aptoula]{sahinbandbased}
I.~Sahin, A.~Erturk, and E.~Aptoula.
\newblock Band-based interpretability with shap for hyperspectral
  classification.
\newblock In \emph{2023 31st Signal Processing and Communications Applications
  Conference (SIU)}, pages 1--4, Istanbul, Turkiye, 2023.

\bibitem[Selvaraju et~al.(2017)Selvaraju, Cogswell, Das, Vedantam, Parikh, and
  Batra]{selvaraju2017grad}
R.~R. Selvaraju, M.~Cogswell, A.~Das, R.~Vedantam, D.~Parikh, and D.~Batra.
\newblock Grad-cam: Visual explanations from deep networks via gradient-based
  localization.
\newblock In \emph{Proceedings of the IEEE international conference on computer
  vision}, pages 618--626, 2017.

\bibitem[Shapley(1953)]{Shapley1953}
L.~S. Shapley.
\newblock A value for n-person games.
\newblock In H.~W. Kuhn and A.~W. Tucker, editors, \emph{Contributions to the
  Theory of Games II}, pages 307--317. Princeton University Press, Princeton,
  1953.

\bibitem[Shen et~al.(2021)Shen, Liu, Wu, Yang, and Xiao]{shen2021admm}
D.~Shen, J.~Liu, Z.~Wu, J.~Yang, and L.~Xiao.
\newblock Admm-hfnet: A matrix decomposition-based deep approach for
  hyperspectral image fusion.
\newblock \emph{IEEE Transactions on Geoscience and Remote Sensing},
  60:\penalty0 1--17, 2021.

\bibitem[Shen et~al.(2022)Shen, Jiang, Li, Zhou, Yuan, and
  Zhang]{shen2022coupling}
H.~Shen, M.~Jiang, J.~Li, C.~Zhou, Q.~Yuan, and L.~Zhang.
\newblock Coupling model-and data-driven methods for remote sensing image
  restoration and fusion: Improving physical interpretability.
\newblock \emph{IEEE Geoscience and Remote Sensing Magazine}, 10\penalty0
  (2):\penalty0 231--249, 2022.

\bibitem[Shi et~al.(2022)Shi, Gu, Gao, Liu, and Chen]{shi2022optical}
M.~Shi, Z.~Gu, Y.~Gao, X.~Liu, and L.~Chen.
\newblock Optical remote sensing image deblurring based on deep unfolding.
\newblock In \emph{IGARSS 2022-2022 IEEE International Geoscience and Remote
  Sensing Symposium}, pages 3295--3298, 2022.

\bibitem[Shrikumar et~al.(2017)Shrikumar, Greenside, and
  Kundaje]{Shrikumar2016}
A.~Shrikumar, P.~Greenside, and A.~Kundaje.
\newblock Learning important features through propagating activation
  differences.
\newblock In \emph{Proceedings of the 34th International Conference on Machine
  Learning}, volume~70, pages 3145–--3153, 2017.

\bibitem[Son and Stratoulias(2022)]{son2022sentinel}
R.~Son and D.~Stratoulias.
\newblock Sentinel-5p based estimation of pm 2.5 concentrations across thailand
  using tabnet.
\newblock In \emph{Proceedings of the IEEE International Geoscience and Remote
  Sensing Symposium (IGARSS)}, pages 6618--6621. IEEE, 2022.

\bibitem[Springenberg et~al.(2014)Springenberg, Dosovitskiy, Brox, and
  Riedmiller]{springenberg2014striving}
J.~T. Springenberg, A.~Dosovitskiy, T.~Brox, and M.~Riedmiller.
\newblock Striving for simplicity: The all convolutional net.
\newblock \emph{arXiv preprint arXiv:1412.6806}, 2014.

\bibitem[Springenberg et~al.(2015)Springenberg, Dosovitskiy, Brox, and
  Riedmiller]{Springenberg2015}
J.~T. Springenberg, A.~Dosovitskiy, T.~Brox, and M.~A. Riedmiller.
\newblock Striving for simplicity: The all convolutional net.
\newblock In Y.~Bengio and Y.~LeCun, editors, \emph{3rd International
  Conference on Learning Representations {ICLR}}, pages 1--14, San Diego, USA,
  2015.

\bibitem[Stadtler et~al.(2022)Stadtler, Betancourt, and
  Roscher]{stadtler2022explainable}
S.~Stadtler, C.~Betancourt, and R.~Roscher.
\newblock Explainable machine learning reveals capabilities, redundancy, and
  limitations of a geospatial air quality benchmark dataset.
\newblock \emph{Machine learning and knowledge extraction}, 4\penalty0
  (1):\penalty0 150--171, 2022.

\bibitem[Sun et~al.(2021)Sun, Liu, Yang, Xiao, and Wu]{sun2021deep}
Y.~Sun, J.~Liu, J.~Yang, Z.~Xiao, and Z.~Wu.
\newblock A deep image prior-based interpretable network for hyperspectral
  image fusion.
\newblock \emph{Remote Sensing Letters}, 12\penalty0 (12):\penalty0 1250--1259,
  2021.

\bibitem[Sundararajan et~al.(2017)Sundararajan, Taly, and
  Yan]{Sundararajan2017}
M.~Sundararajan, A.~Taly, and Q.~Yan.
\newblock Axiomatic attribution for deep networks.
\newblock In \emph{International conference on machine learning}, pages
  3319--3328, 2017.

\bibitem[Swartout and Moore(1993)]{swartout1993explanation}
W.~R. Swartout and J.~D. Moore.
\newblock {Explanation in second generation expert systems}.
\newblock In \emph{Second generation expert systems}, pages 543--585. Springer,
  1993.

\bibitem[Tan et~al.(2022)Tan, Xiao, Zhu, Wan, Wang, and Li]{tan2022transformer}
X.~Tan, Z.~Xiao, J.~Zhu, Q.~Wan, K.~Wang, and D.~Li.
\newblock Transformer-driven semantic relation inference for multilabel
  classification of high-resolution remote sensing images.
\newblock \emph{IEEE Journal of Selected Topics in Applied Earth Observations
  and Remote Sensing}, 15:\penalty0 1884--1901, 2022.

\bibitem[Taskin(2022)]{taskin2022model}
G.~Taskin.
\newblock A model distillation approach for explaining black-box models for
  hyperspectral image classification.
\newblock In \emph{Proceedings of the {IEEE} International Geoscience and
  Remote Sensing Symposium (IGARSS)}, pages 3592--3595, 2022.

\bibitem[Teach and Shortliffe(1981)]{teach1981analysis}
R.~L. Teach and E.~H. Shortliffe.
\newblock {An analysis of physician attitudes regarding computer-based clinical
  consultation systems}.
\newblock \emph{Computers and Biomedical Research}, 14\penalty0 (6):\penalty0
  542--558, 1981.

\bibitem[Temenos et~al.(2022)Temenos, Tzortzis, Kaselimi, Rallis, Doulamis, and
  Doulamis]{temenos2022novel}
A.~Temenos, I.~N. Tzortzis, M.~Kaselimi, I.~Rallis, A.~Doulamis, and
  N.~Doulamis.
\newblock Novel insights in spatial epidemiology utilizing explainable ai (xai)
  and remote sensing.
\newblock \emph{Remote Sensing}, 14\penalty0 (13):\penalty0 3074, 2022.

\bibitem[Temenos et~al.(2023)Temenos, Temenos, Kaselimi, Doulamis, and
  Doulamis]{temenos2023interpretable}
A.~Temenos, N.~Temenos, M.~Kaselimi, A.~Doulamis, and N.~Doulamis.
\newblock Interpretable deep learning framework for land use and land cover
  classification in remote sensing using shap.
\newblock \emph{IEEE Geoscience and Remote Sensing Letters}, 20:\penalty0 1--5,
  2023.

\bibitem[Tian et~al.(2021)Tian, Li, Wang, and Ma]{tian2021vp}
X.~Tian, K.~Li, Z.~Wang, and J.~Ma.
\newblock Vp-net: An interpretable deep network for variational pansharpening.
\newblock \emph{IEEE Transactions on Geoscience and Remote Sensing},
  60:\penalty0 1--16, 2021.

\bibitem[Tuia et~al.(2021)Tuia, Roscher, Wegner, Jacobs, Zhu, and
  Camps-Valls]{tuia2021toward}
D.~Tuia, R.~Roscher, J.~D. Wegner, N.~Jacobs, X.~Zhu, and G.~Camps-Valls.
\newblock {Toward a collective agenda on {AI} for {Earth} science data
  analysis}.
\newblock \emph{IEEE Geoscience and Remote Sensing Magazine}, 9\penalty0
  (2):\penalty0 88--104, 2021.

\bibitem[Vald{\'e}s and Pou(2021)]{valdes2021machine}
J.~J. Vald{\'e}s and A.~Pou.
\newblock A machine learning-explainable ai approach to tropospheric dynamics
  analysis using water vapor meteosat images.
\newblock In \emph{2021 IEEE Symposium Series on Computational Intelligence
  (SSCI)}, pages 1--8. IEEE, 2021.

\bibitem[Vasu et~al.(2018)Vasu, Rahman, and Savakis]{vasu2018aerial}
B.~Vasu, F.~U. Rahman, and A.~Savakis.
\newblock Aerial-cam: Salient structures and textures in network class
  activation maps of aerial imagery.
\newblock In \emph{2018 IEEE 13th Image, Video, and Multidimensional Signal
  Processing Workshop (IVMSP)}, pages 1--5. IEEE, 2018.

\bibitem[Verma et~al.(2021)Verma, Gupta, Tolani, and
  Kaushal]{verma2021explainable}
M.~Verma, N.~Gupta, B.~Tolani, and R.~Kaushal.
\newblock Explainable custom {CNN} architecture for land use classification
  using satellite images.
\newblock In \emph{2021 Sixth International Conference on Image Information
  Processing (ICIIP)}, volume~6, pages 304--309. IEEE, 2021.

\bibitem[Walker et~al.(2021)Walker, Peeples, Dale, Keller, and
  Zare]{walker2021explainable}
S.~Walker, J.~Peeples, J.~Dale, J.~Keller, and A.~Zare.
\newblock Explainable systematic analysis for synthetic aperture sonar imagery.
\newblock In \emph{Proceedings of the IEEE International Geoscience and Remote
  Sensing Symposium (IGARSS)}, pages 2835--2838. IEEE, 2021.

\bibitem[Wang et~al.(2019{\natexlab{a}})Wang, Chen, Xu, Liu, Loy, and
  Lin]{wang2019carafe}
J.~Wang, K.~Chen, R.~Xu, Z.~Liu, C.~C. Loy, and D.~Lin.
\newblock Carafe: Content-aware reassembly of features.
\newblock In \emph{Proceedings of the IEEE/CVF international conference on
  computer vision}, pages 3007--3016, 2019{\natexlab{a}}.

\bibitem[Wang et~al.(2022{\natexlab{a}})Wang, Shao, Huang, Lu, and
  Zhang]{wang2022deep}
J.~Wang, Z.~Shao, X.~Huang, T.~Lu, and R.~Zhang.
\newblock A deep unfolding method for satellite super resolution.
\newblock \emph{IEEE Transactions on Computational Imaging},
  2022{\natexlab{a}}.

\bibitem[Wang et~al.(2019{\natexlab{b}})Wang, Zeng, Huang, Ding, and
  Paisley]{wang2019deep}
W.~Wang, W.~Zeng, Y.~Huang, X.~Ding, and J.~Paisley.
\newblock Deep blind hyperspectral image fusion.
\newblock In \emph{Proceedings of the IEEE/CVF International Conference on
  Computer Vision}, pages 4150--4159, 2019{\natexlab{b}}.

\bibitem[Wang et~al.(2021)Wang, Fu, Zeng, Sun, Zhan, Huang, and
  Ding]{wang2021enhanced}
W.~Wang, X.~Fu, W.~Zeng, L.~Sun, R.~Zhan, Y.~Huang, and X.~Ding.
\newblock Enhanced deep blind hyperspectral image fusion.
\newblock \emph{IEEE transactions on neural networks and learning systems},
  2021.

\bibitem[Wang et~al.(2022{\natexlab{b}})Wang, Abliz, Ma, Liu, Kurban, Halik,
  Pietik{\"a}inen, and Wang]{wang2022hyperspectral}
Y.~Wang, A.~Abliz, H.~Ma, L.~Liu, A.~Kurban, {\"U}.~Halik, M.~Pietik{\"a}inen,
  and W.~Wang.
\newblock Hyperspectral estimation of soil copper concentration based on
  improved tabnet model in the eastern junggar coalfield.
\newblock \emph{IEEE Transactions on Geoscience and Remote Sensing},
  60:\penalty0 1--20, 2022{\natexlab{b}}.

\bibitem[Wolanin et~al.(2020)Wolanin, Mateo-Garc{\'\i}a, Camps-Valls,
  G{\'o}mez-Chova, Meroni, Duveiller, Liangzhi, and
  Guanter]{wolanin2020estimating}
A.~Wolanin, G.~Mateo-Garc{\'\i}a, G.~Camps-Valls, L.~G{\'o}mez-Chova,
  M.~Meroni, G.~Duveiller, Y.~Liangzhi, and L.~Guanter.
\newblock Estimating and understanding crop yields with explainable deep
  learning in the indian wheat belt.
\newblock \emph{Environmental research letters}, 15\penalty0 (2):\penalty0
  024019, 2020.

\bibitem[Xia et~al.(2018)Xia, Bai, Ding, Zhu, Belongie, Luo, Datcu, Pelillo,
  and Zhang]{xia2018dota}
G.-S. Xia, X.~Bai, J.~Ding, Z.~Zhu, S.~Belongie, J.~Luo, M.~Datcu, M.~Pelillo,
  and L.~Zhang.
\newblock Dota: A large-scale dataset for object detection in aerial images.
\newblock In \emph{Proceedings of the IEEE conference on computer vision and
  pattern recognition}, pages 3974--3983, 2018.

\bibitem[Xie et~al.(2021)Xie, Meng, Li, Li, Yu, Sun, Song, and
  Xu]{xie2021visual}
P.~Xie, F.~Meng, B.~Li, Y.~Li, Z.~Yu, H.~Sun, T.~Song, and D.~Xu.
\newblock Visual prediction of tropical cyclones with deep convolutional
  generative adversarial networks.
\newblock In \emph{Proceedings of the IEEE International Geoscience and Remote
  Sensing Symposium (IGARSS)}, pages 8297--8300. IEEE, 2021.

\bibitem[Xie et~al.(2019)Xie, Zhou, Zhao, Meng, Zuo, and
  Xu]{xie2019multispectral}
Q.~Xie, M.~Zhou, Q.~Zhao, D.~Meng, W.~Zuo, and Z.~Xu.
\newblock Multispectral and hyperspectral image fusion by ms/hs fusion net.
\newblock In \emph{Proceedings of the IEEE/CVF Conference on Computer Vision
  and Pattern Recognition}, pages 1585--1594, 2019.

\bibitem[Xie et~al.(2020)Xie, Zhou, Zhao, Xu, and Meng]{xie2020mhf}
Q.~Xie, M.~Zhou, Q.~Zhao, Z.~Xu, and D.~Meng.
\newblock Mhf-net: An interpretable deep network for multispectral and
  hyperspectral image fusion.
\newblock \emph{IEEE Transactions on Pattern Analysis and Machine
  Intelligence}, 2020.

\bibitem[Xiong et~al.(2021{\natexlab{a}})Xiong, Zhou, Tao, Lu, and
  Qian]{xiong2021snmf}
F.~Xiong, J.~Zhou, S.~Tao, J.~Lu, and Y.~Qian.
\newblock Snmf-net: Learning a deep alternating neural network for
  hyperspectral unmixing.
\newblock \emph{IEEE Transactions on Geoscience and Remote Sensing},
  60:\penalty0 1--16, 2021{\natexlab{a}}.

\bibitem[Xiong et~al.(2021{\natexlab{b}})Xiong, Zhou, Ye, Lu, and
  Qian]{xiong2021nmf}
F.~Xiong, J.~Zhou, M.~Ye, J.~Lu, and Y.~Qian.
\newblock Nmf-sae: An interpretable sparse autoencoder for hyperspectral
  unmixing.
\newblock In \emph{ICASSP 2021-2021 IEEE International Conference on Acoustics,
  Speech and Signal Processing (ICASSP)}, pages 1865--1869. IEEE,
  2021{\natexlab{b}}.

\bibitem[Xiong et~al.(2022)Xiong, Xiong, and Cui]{xiong2022explainable}
W.~Xiong, Z.~Xiong, and Y.~Cui.
\newblock An explainable attention network for fine-grained ship classification
  using remote-sensing images.
\newblock \emph{IEEE Transactions on Geoscience and Remote Sensing},
  60:\penalty0 1--14, 2022.

\bibitem[Xu et~al.(2021)Xu, Zhang, Zhao, Sun, Liu, and Zhang]{xu2021deep}
S.~Xu, J.~Zhang, Z.~Zhao, K.~Sun, J.~Liu, and C.~Zhang.
\newblock Deep gradient projection networks for pan-sharpening.
\newblock In \emph{Proceedings of the IEEE/CVF Conference on Computer Vision
  and Pattern Recognition}, pages 1366--1375, 2021.

\bibitem[Yang et~al.(2022)Yang, Zhou, Yan, Liu, Fu, and Wang]{yang2022memory}
G.~Yang, M.~Zhou, K.~Yan, A.~Liu, X.~Fu, and F.~Wang.
\newblock Memory-augmented deep conditional unfolding network for
  pan-sharpening.
\newblock In \emph{Proceedings of the IEEE/CVF Conference on Computer Vision
  and Pattern Recognition}, pages 1788--1797, 2022.

\bibitem[Yang et~al.(2021)Yang, Xiao, Zhao, and Chan]{yang2021variational}
J.~Yang, L.~Xiao, Y.-Q. Zhao, and J.~C.-W. Chan.
\newblock Variational regularization network with attentive deep prior for
  hyperspectral--multispectral image fusion.
\newblock \emph{IEEE Transactions on Geoscience and Remote Sensing},
  60:\penalty0 1--17, 2021.

\bibitem[Yang et~al.(2019)Yang, Xu, Xu, Ding, and Pu]{yang2019class}
R.~Yang, X.~Xu, Z.~Xu, C.~Ding, and F.~Pu.
\newblock A class activation mapping guided adversarial training method for
  land-use classification and object detection.
\newblock In \emph{Proceedings of the IEEE International Geoscience and Remote
  Sensing Symposium}, pages 9474--9477, 2019.

\bibitem[Yasuma et~al.(2010)Yasuma, Mitsunaga, Iso, and
  Nayar]{yasuma2010generalized}
F.~Yasuma, T.~Mitsunaga, D.~Iso, and S.~K. Nayar.
\newblock Generalized assorted pixel camera: postcapture control of resolution,
  dynamic range, and spectrum.
\newblock \emph{IEEE transactions on image processing}, 19\penalty0
  (9):\penalty0 2241--2253, 2010.

\bibitem[Ye and Johnson(1995)]{ye1995impact}
L.~R. Ye and P.~E. Johnson.
\newblock The impact of explanation facilities on user acceptance of expert
  systems advice.
\newblock \emph{Mis Quarterly}, pages 157--172, 1995.

\bibitem[Yin(2021)]{yin2021pscsc}
H.~Yin.
\newblock Pscsc-net: A deep coupled convolutional sparse coding network for
  pansharpening.
\newblock \emph{IEEE Transactions on Geoscience and Remote Sensing},
  60:\penalty0 1--16, 2021.

\bibitem[Zhang et~al.(2018{\natexlab{a}})Zhang, Bargal, Lin, Brandt, Shen, and
  Sclaroff]{zhang2018top}
J.~Zhang, S.~A. Bargal, Z.~Lin, J.~Brandt, X.~Shen, and S.~Sclaroff.
\newblock Top-down neural attention by excitation backprop.
\newblock \emph{International Journal of Computer Vision}, 126\penalty0
  (10):\penalty0 1084--1102, 2018{\natexlab{a}}.

\bibitem[Zhang et~al.(2020{\natexlab{a}})Zhang, Nie, Wei, Li, and
  Zhang]{zhang2020deep}
L.~Zhang, J.~Nie, W.~Wei, Y.~Li, and Y.~Zhang.
\newblock Deep blind hyperspectral image super-resolution.
\newblock \emph{IEEE Transactions on Neural Networks and Learning Systems},
  32\penalty0 (6):\penalty0 2388--2400, 2020{\natexlab{a}}.

\bibitem[Zhang et~al.(2018{\natexlab{b}})Zhang, Wu, and
  Zhu]{zhang2018interpretable}
Q.~Zhang, Y.~N. Wu, and S.-C. Zhu.
\newblock Interpretable convolutional neural networks.
\newblock In \emph{Proceedings of the IEEE conference on computer vision and
  pattern recognition}, pages 8827--8836, 2018{\natexlab{b}}.

\bibitem[Zhang et~al.(2020{\natexlab{b}})Zhang, Huang, Wang, and
  Li]{zhang2020ssr}
X.~Zhang, W.~Huang, Q.~Wang, and X.~Li.
\newblock Ssr-net: Spatial--spectral reconstruction network for hyperspectral
  and multispectral image fusion.
\newblock \emph{IEEE Transactions on Geoscience and Remote Sensing},
  59\penalty0 (7):\penalty0 5953--5965, 2020{\natexlab{b}}.

\bibitem[Zhang et~al.(2020{\natexlab{c}})Zhang, Lv, Yao, Xiong, and
  Fu]{zhang2020new}
X.~Zhang, Y.~Lv, L.~Yao, W.~Xiong, and C.~Fu.
\newblock A new benchmark and an attribute-guided multilevel feature
  representation network for fine-grained ship classification in optical remote
  sensing images.
\newblock \emph{IEEE Journal of Selected Topics in Applied Earth Observations
  and Remote Sensing}, 13:\penalty0 1271--1285, 2020{\natexlab{c}}.

\bibitem[Zhao et~al.(2019)Zhao, Datcu, Zhang, Xiong, and
  Yu]{zhao2019contrastive}
J.~Zhao, M.~Datcu, Z.~Zhang, H.~Xiong, and W.~Yu.
\newblock Contrastive-regulated {CNN} in the complex domain: A method to learn
  physical scattering signatures from flexible polsar images.
\newblock \emph{IEEE Transactions on Geoscience and Remote Sensing},
  57\penalty0 (12):\penalty0 10116--10135, 2019.

\bibitem[Zhao et~al.(2022)Zhao, Ma, Lyu, and Chen]{qi2022embedded}
Q.~Zhao, Y.~Ma, S.~Lyu, and L.~Chen.
\newblock Embedded self-distillation in compact multibranch ensemble network
  for remote sensing scene classification.
\newblock \emph{IEEE Transactions on Geoscience and Remote Sensing},
  60:\penalty0 1--15, 2022.

\bibitem[Zheng et~al.(2020)Zheng, Li, Li, Guo, Wu, and
  Chanussot]{zheng2020hyperspectral}
Y.~Zheng, J.~Li, Y.~Li, J.~Guo, X.~Wu, and J.~Chanussot.
\newblock Hyperspectral pansharpening using deep prior and dual attention
  residual network.
\newblock \emph{IEEE transactions on geoscience and remote sensing},
  58\penalty0 (11):\penalty0 8059--8076, 2020.

\bibitem[Zhou et~al.(2016)Zhou, Khosla, Lapedriza, Oliva, and
  Torralba]{zhou2016learning}
B.~Zhou, A.~Khosla, A.~Lapedriza, A.~Oliva, and A.~Torralba.
\newblock Learning deep features for discriminative localization.
\newblock In \emph{Proceedings of the IEEE conference on computer vision and
  pattern recognition}, pages 2921--2929, 2016.

\bibitem[Zhu et~al.(2015)Zhu, Chen, Dai, Fu, Ye, and Jiao]{zhu2015orientation}
H.~Zhu, X.~Chen, W.~Dai, K.~Fu, Q.~Ye, and J.~Jiao.
\newblock Orientation robust object detection in aerial images using deep
  convolutional neural network.
\newblock In \emph{2015 IEEE International Conference on Image Processing
  (ICIP)}, pages 3735--3739. IEEE, 2015.

\bibitem[Zhu et~al.(2017)Zhu, Tuia, Mou, Xia, Zhang, Xu, and
  Fraundorfer]{Zhu2017}
X.~X. Zhu, D.~Tuia, L.~Mou, G.-S. Xia, L.~Zhang, F.~Xu, and F.~Fraundorfer.
\newblock Deep learning in remote sensing: A comprehensive review and list of
  resources.
\newblock \emph{IEEE Geoscience and Remote Sensing Magazine}, 5\penalty0
  (4):\penalty0 8--36, 2017.

\end{thebibliography}






\end{document}